\pdfoutput=1
\documentclass[11pt]{article}

\usepackage[]{acl}

\usepackage{times}
\usepackage{latexsym}

\usepackage[T1]{fontenc}
\usepackage[utf8]{inputenc}

\usepackage{microtype}

\usepackage{inconsolata}

\usepackage{graphicx}

\usepackage{multirow}
\usepackage{amsmath}
\usepackage{algorithm} 
\usepackage{algorithmic}
\usepackage{subfig}
\usepackage{booktabs}

\title{SH2: Self-Highlighted Hesitation Helps You Decode More Truthfully}

\author{Jushi Kai \quad Tianhang Zhang \quad Hai Hu \quad Zhouhan Lin\Thanks{~Corresponding author} \\
\textsuperscript{} Shanghai Jiao Tong University \\
\textsuperscript{} json.kai@sjtu.edu.cn \quad lin.zhouhan@gmail.com \\}

\begin{document}
\maketitle
\begin{abstract}
Large language models (LLMs) demonstrate great performance in text generation. However, LLMs are still suffering from hallucinations. In this work, we propose an inference-time method, \textbf{S}elf-\textbf{H}ighlighted \textbf{H}esitation (\textbf{SH2}), to help LLMs decode more truthfully. SH2 is based on a simple fact rooted in information theory that for an LLM, the tokens predicted with lower probabilities are prone to be more informative than others. Our analysis shows that these low-confidence tokens are more likely to be closely related to factual information, such as nouns, proper nouns, and adjectives. Therefore, we propose to ``highlight'' the factual information by selecting key tokens with the lowest probabilities and concatenating them to the original context, thus forcing the model to repeatedly read and hesitate on these tokens before generation. During decoding, we also adopt contrastive decoding to emphasize the difference in output probabilities brought by the hesitation. 
Experimental results demonstrate that our SH2, requiring no additional data or models, can effectively help LLMs elicit factual knowledge and distinguish hallucinated contexts by themselves. Significant and consistent improvements are achieved by SH2 for LLaMA-7b, LLaMA2-7b and Mistral-7b on various hallucination tasks.\footnote{We release our code at \url{https://github.com/LUMIA-Group/SH2}.} %\footnote{We will release our code for reproducibility later.}

\end{abstract}

\section{Introduction}

Depending on massive training corpora, large language models (LLMs) have made tremendous progress in natural language understanding and text generation (\citealp{llama}, \citealp{mistral}, \citealp{gpt4}). However, during reasoning and generation, LLMs could suffer from hallucinations and generate non-factual answers (\citealp{siren}).

% The first ones focus on the misinformation from the training data. Crawled from the internet, a lot of outdated, biased and untruthful information is mixed in with the training data and stuffed into models. Although some researchers have been working on clearing these falsehoods and constructing datasets with higher quality (\citealp{moss}, \citealp{llama2}, \citealp{lima}), it seems impossible to guarantee zero error in the training data.

To clear these falsehoods, some researchers construct datasets with higher quality and train LLMs to respond in the correct form (\citealp{alpaca}, \citealp{lima}). But in the domain not included in the training, LLMs assign similar probabilities to correct and wrong choices since they do not have enough relevant knowledge to distinguish them.

To fill the gap of knowledge, retrieval augmentation methods \cite{check, rarr, critic} leverage external knowledge bases and tools to correct the output of LLMs. Although they provide additional information for LLMs, it still seems impossible to guarantee zero error in the external knowledge. Other researchers propose decoding reformulation methods \cite{iti, cd-reasoning, dola} to address hallucinations. However, most of them rely on external data with human labor or larger models to rescale probability distribution during decoding.

Recent works \cite{cot,pause-training} have demonstrated that a few more computation steps for LLMs can make a difference. \citet{cot} propose chain-of-thought (COT) prompting to elicit intermediate reasoning steps of LLMs. These intermediate steps can lead to better answers. However, elaborate prompts need to be prepared in advance. \citet{pause-training} append learnable pause tokens to the input prefix and train models from scratch. These delays introduced by pause tokens provide the model with more computation steps to generate better answers.% Our work follows the idea of using extra computation steps during decoding.

In this paper, we seek to leverage the LLM itself to highlight informative tokens and digest them as input during the hesitation steps to elicit truthful knowledge inside LLMs. We propose a simple yet effective method, \textbf{S}elf-\textbf{H}ighlighted \textbf{H}esitation (\textbf{SH2}), to help LLMs decode more truthfully. SH2 introduces hesitations to give LLMs more time to understand contexts and answer questions. For LLMs, the tokens assigned with lower probabilities are harder to predict, while more likely to be informative. \textit{LLMs can select these key tokens by themselves from the input and hesitate on these highlighted tokens.} We calculate the difference brought by highlighted tokens through contrastive decoding (\citealp{cd}) and integrate it into the output probability. Experiments on multiple tasks demonstrate that such a difference could elicit factual knowledge inside the model and successfully mitigate hallucinations. Unlike other methods, our method does not leverage any other external tools or data. Additionally, it can be directly deployed during inference with no more training.

\section{Related Work}

\subsection{Hallucination Mitigation}
Recent works to mitigate the hallucination of LLMs can be summed up into three categories.

% One is mitigating during the training process. 

\paragraph{Supervised Fine-Tuning}
Many researchers pay attention to the curation of the training data and attempt to mitigate hallucinations through additional fine-tuning. Alpaca \cite{alpaca} have collected 52K instruction-following data of massive tasks and fine-tuned the LLaMA-7b model \cite{llama}. Such an instruction tuning process is also known as supervised fine-tuning (SFT). \citet{lima} construct 1000 SFT samples with human labor for alignment and suggest that almost all knowledge in LLMs has been learned during pretraining. Moreover, \citet{alpagasus} leverage ChatGPT to automatically select high-quality data from Alpaca. These training approaches have high requirements for computational resources.

% Since training approaches have high requirements for computational resources, many inference-time approaches are proposed to circumvent hallucinations. These methods can also be divided into retrieval augmentation and decoding reformulation.

\paragraph{Retrieval Augmentation}
Retrieval augmentation approaches resort to external knowledge bases and tools to help correct hallucinations. Additional information is retrieved to provide relevant knowledge for LLMs and support their generation. \citet{check} and \citet{rarr} leverage search engines to attribute and refine the output of language models. 
% \citet{check} leverage search engines and external databases to consolidate evidence for the query fed into LLMs and revise their responses. Besides researching and revising, \citet{rarr} introduce an agreement model and an edit model to help refine the output of language models. 
\citet{critic} enable multiple tools to correct responses of LLMs autonomously during the interaction with external tools.

\paragraph{Decoding Reformulation}
These approaches work on reformulating the probability distribution of outputs. \citet{iti} introduce Inference-Time Intervention (ITI) to locate truthful directions of TruthfulQA \cite{truthfulqa} and shift model activations toward truthfulness during inference. However, they need the data of TruthfulQA to train a domain-specific classifier for each attention head. \citet{cd} propose Contrastive Decoding (CD) to capture the likelihood difference between large and small models. The difference signals which input texts should be preferred during decoding. \citet{cd-reasoning} utilize CD to improve reasoning quality for LLMs. Furthermore, DoLa \cite{dola} use the last layer as the expert model and the premature layer as the amateur, and contrast prediction probabilities between them. Different from these model-based CD strategies, our proposed method diverges by contrasting probabilities from a data-based perspective, offering a novel angle for decoding reformulation.

\subsection{More Computations for Decoding}
The idea of using extra decoding steps when predicting hard tokens can be dated back to Adaptive Computation Time proposed by \citet{act}. \citet{autoprompt} introduce AUTOPROMPT to combine original inputs with trigger tokens to elicit knowledge from pretrained models. \citet{rationale} investigate that additional steps to generate rationales could lead to a more faithful model. More recently, \citet{pause-training} have also demonstrated that for LLMs, inserting extra computation steps to allow the model to pause before generation can enhance the performance on question answering and reasoning tasks. Nevertheless, their method only works when the model is both pre-trained and finetuned with pauses.

\begin{figure*} [ht]
    \centering
    \includegraphics[width=1.02\textwidth]
    {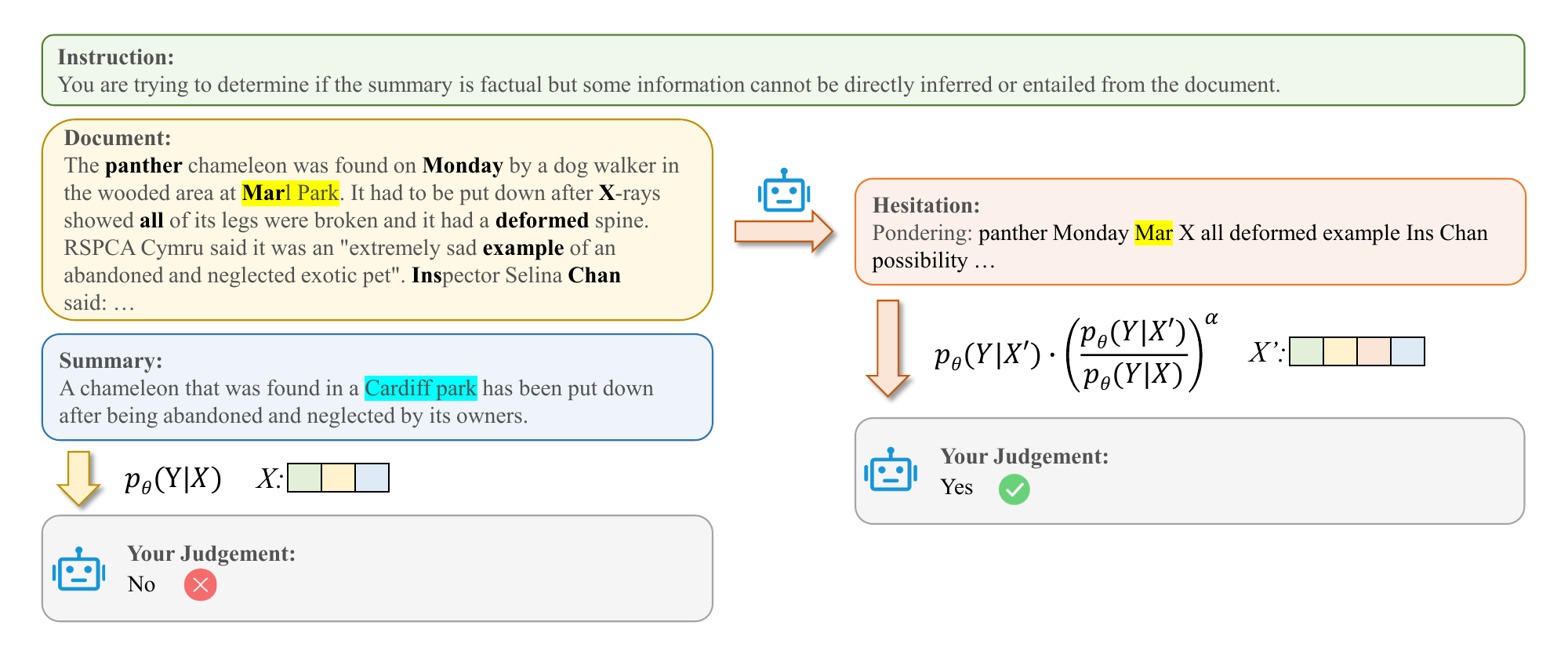}
    \caption{The pipeline to construct and leverage our Self-Highlighted Hesitation. The original input \textit{X} consists of the instruction, document and summary. The hesitation of key tokens is appended to the document in the hesitated input \textit{X'}.}
    \label{fig-framework}
\end{figure*}

\section{Self-Highlighted Hesitation}

The illustration of our Self-Highlighted Hesitation is shown in Figure \ref{fig-framework}. For the original inference procedure, the instruction, document, and summary are directly fed into the LLM to generate the judgment. The LLM could be easily confused by the hallucinated context.

For our SH2, we underline the key tokens of the input document by the prediction likelihood. These tokens are listed as a hesitation and appended to the document. The prediction probability is scaled with the difference of confidence between the hesitated input \textit{X'} and the original input \textit{X}. Our method can effectively help LLMs identify the hallucinated context. We will elaborate our method in the following subsections.

\subsection{Key Tokens}

We select key tokens based on the prediction probability given by the LLM. The decoding procedure of a language model $\theta$ can be formalized as:
\begin{equation}
    \hat{p}(x_t) = p_\theta(x_t|x_{<t})
\end{equation}
where $x$ is the context, $t$ denotes the current predicting position of $x$ and $p_\theta$ gives the prediction probability of the token $x_t$ by the model $\theta$. We obtain the probability of generating $x_t$ by feeding previous tokens to the model.

The probability measures the confidence of the language model for each token given the previous context. It represents how simple it is to infer the token from the previous context by the model. The tokens with the lowest probabilities bring the most semantic information and are the hardest to predict. We regard these tokens as \textbf{Key Tokens}. They deserve more attention from the language model and help comprehend the whole context.

% We sample $n=1000$ documents from the summarization track of HaluEval \cite{HaluEval}, which is a dataset collected from CNN/Daily Mail \cite{cnndm}.

\begin{figure*} [t]
    \centering
    \includegraphics[width=1.01\linewidth]
    {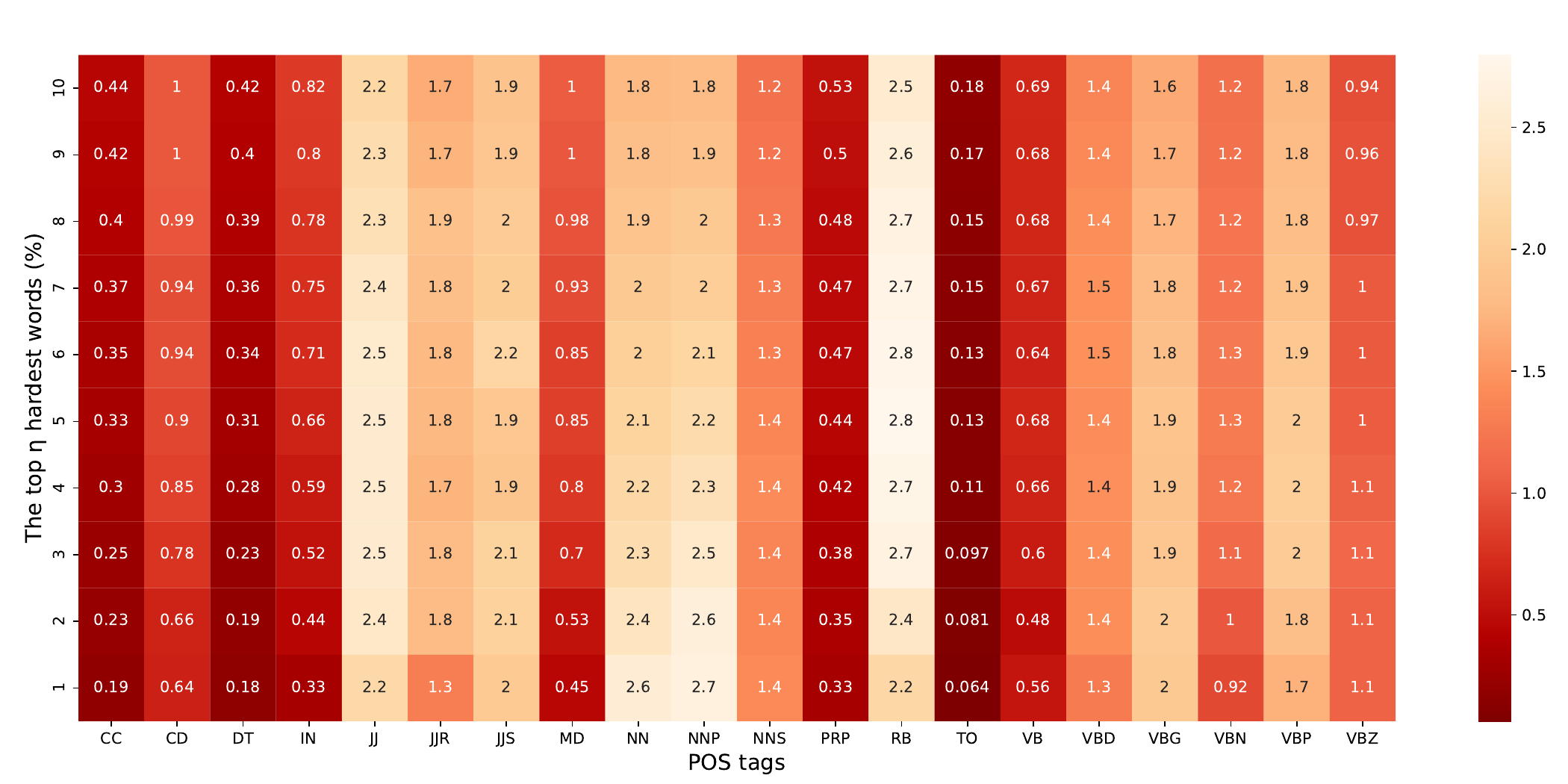}
    \caption{The heatmap to show the normalized top-$\eta$ recall for the top 20 most frequent POS tags. The light color and the high value indicates that these POS tags occupy high proportions in the hardest part. 1000 documents are sampled from the summarization track of HaluEval \cite{HaluEval}, which is a dataset collected from CNN/Daily Mail \cite{cnndm}. We extract the hardest words that contain key tokens from these documents with the proportion of $\eta$ ranging from 1\% to 10\% by LLaMA2-7b \cite{llama2}. }
    \label{fig-parsing}
\end{figure*}

\subsection{Relation between Key Tokens and Factual Information}
\label{sec:relation}

To illustrate that key tokens selected in this fashion are closely related to the factual knowledge, we analyze the \textbf{Normalized Top-$\eta$ Recall} for POS (part-of-speech) tags from the perspective of grammar. It measures the percentage of POS tags in the document that appear in the hardest part.

For the document $X_i$, the number of words with POS tag $z_k$ is:
\begin{equation}
    N(X_i, z_k) = \#\{POS(X_{i})=z_k\}
\end{equation}
where $POS(\cdot)$ is the function to derive POS tags and $\#\{POS(X_{i})=z_k\}$ is to count how many words in $X_i$ have the POS tag of $z_k$.

For a dataset with $m$ documents, the normalized recall of POS tag $z_k$ can be calculated by:
\begin{equation}
\label{eq:delta}
    \Delta_\eta(z_k) = \frac{\sum_{1 \leq i \leq m} N(T(X_i, \eta), z_k)}{\eta\cdot \sum_{1 \leq i \leq m} N(X_i,z_k)}
\end{equation}
where $T(X_i, \eta)$ is the set of words that are among the lowest $\eta$ portion of probability predicted by the language model in $X_i$. The numerator measures the frequency of the POS tag $z_k$ in the subset of the lowest probability. The $\eta$ in the denominator is to normalize the scale of $\Delta_\eta(z_k)$ with different $\eta$.

$\Delta_\eta(z_k)$ is the frequency difference between subsets and documents. It measures how hard words with the POS tag $z_k$ are for LLMs to predict. Figure \ref{fig-parsing} demonstrates the normalized top-$\eta$ recall of the most frequent POS tags for the summarization track of HaluEval \cite{HaluEval}.

Larger $\Delta_\eta(z_k)$ means that $z_k$ is more concentrated in the hardest part of documents. For example, although there are 10 times as many prepositions (IN) as superlative adjectives (JJS) in the set of words with the lowest 1\% ($\eta=1$) portion of probability, the recall of IN is even smaller. This is because IN is more frequent naturally. There are 60 times as many IN as JJS throughout documents, but IN is less concentrated in the hardest part.% $\Delta_\eta(\text{IN})$ becomes larger as $\eta$ increases.

It can be observed from Figure~\ref{fig-parsing} that \textbf{content words} such as adjectives (JJ), nouns (NN), proper nouns (NNP), adverbs (RB) and conjugated verbs (VBD, VBG, VBP) are more difficult to predict and more concentrated in the hardest part, as the light color in the heatmap indicates. These content words usually contain factual information. On the other hand, \textbf{function words} such as conjunctions (CC), determiners (DT), and prepositions (IN) are less informative and less concentrated in the hardest part.\footnote{The base form of the verbs (VB) seems to be an exception. It is probably because VBs have less information load than conjugated verb forms, as the latter also encode tense, aspect, and other information about the verb.} 
Therefore, picking out key tokens that carry more content, rather than those that serve as grammatical functions, could potentially help the model focus on factual information. 
% Our method can thus be understood as a way to ``highlight'' the content words to LLMs, potentially making them less likely to hallucinate, the effect of which will be evaluated in our experiments. More cases will be given in Appendix \ref{sec:cases}.

% It can be learned from the figure that the distribution difference is positively correlated with factual informativeness. \textbf{Content words}, including nouns (NN), proper nouns (NNP), adjectives (JJ), and adverbs (RB) are the hardest for LLMs to predict. These content words often contain the most semantic information and can help infer the following context. They are probably more relevant to facts and events. While \textbf{function words}, including conjunctions (CC, IN), determiners (DT), personal pronouns (PRP), and \textit{''to''} (TO), are not closely related to factual information and much easier to predict. In addition, carrying more information, conjugations are also hard for LLMs to capture. It can be noted that past tense (VBD), gerund or present participle (VBG), and non-3rd person singular present (VBP) are much more challenging than base form verbs (VB).
% %We also notice that words that appear for the first time in the text are hard to predict. 

% Our findings suggest that key tokens are closely related to factual knowledge. Existing LLMs are not powerful enough to capture important factual information, which could be the primary reason leading to the problem of hallucinations.

\subsection{Construction of Hesitations}

To this end, we attempt to solve hallucinations by highlighting key tokens that are hard to predict. For the text sequence $X=(x_0, x_1, \dots, x_n)$, we can obtain the corresponding probability sequence by the language model:
\begin{equation}
    \hat{P}(X)=(\hat{p}(x_1), \hat{p}(x_2), \dots, \hat{p}(x_n))
\end{equation}

We use the following strategy to select key tokens according to their probability $\hat{P}(X)$ and construct hesitations with these tokens:

First, construct a candidate key-token set by selecting $\eta \cdot n$ tokens with the lowest prediction probabilities of $\hat{P}(X)$, where $\eta \in (0, 1)$ is the sampling proportion we preset for $X$. 
% Then, derive the \textbf{target key-token set} $T(X)$ by randomly drop out $\lambda \cdot \eta \cdot n$ tokens from the candidate set, where $\lambda \in [0, 1]$ is the drop-out rate.

For long documents, to avoid the key-token set getting dominated by one or two POS tags, we introduce the drop-out rate $\lambda \in [0, 1]$ to randomly select key tokens among a larger pool of candidates, whose size is determined by $\eta$. As a result, $(1 - \lambda) \cdot \eta \cdot n$ tokens are retained in the \textbf{target key-token set} $T(X)$ to fill as input during hesitation steps.

Finally, construct the \textbf{Hesitation} $H(X)$ with the target key-token set for the input context: $H(X) = ``Pondering:\space <T(X)>."$, where $``Pondering:\space"$ is the prefix. We keep the tokens' order of $T(X)$ in which they appear in the original text.

With the hesitation text following the original text, the language model can focus more on key tokens and have more time to infer the non-hallucinated answer.

\subsection{Contrastive Decoding on Hesitations}
\label{sec:cd}

Using hesitations as prompting or data augmentation may not be enough. Since we do not introduce extra factual information, the improvement from the hesitated input might be limited. Therefore, we resort to what difference hesitations can bring during decoding. Different from previous work \cite{cd, dola, cd-reasoning} that contrasts between models, we use contrastive decoding (\citealp{cd}) from the perspective of data:
\begin{equation}
\label{eq:cd}
    p_\mathbf{CD}(Y|X) = \text{softmax}\left( \frac{p_\theta(Y|X')}{p_\theta(Y|X)}\right)
\end{equation}
where $Y$ is the output text and $X'$ is the concatenation of the original input text $X$ and its hesitation $H(X)$.

The difference of confidence is traded off with the base probability $p_\theta(Y|X')$:
\begin{equation}
    % \resizebox{1.02\hsize}{!}{
    p_\mathbf{H}(Y|X)=\left\{\begin{array}{cl}
    \hspace{-0.2cm}\text{softmax}(p_\theta(Y|X') \cdot (\frac{p_\theta(Y|X')}{p_\theta(Y|X)})^\alpha),\\
    \quad\quad\quad\quad\quad\quad\quad\quad\quad\quad \alpha \neq 0 \\
    p_\theta(Y|X'), \text { otherwise }
\end{array}\right.
\end{equation}
where $\alpha$ is a hyper-parameter used for scaling the difference of confidence between with and without hesitations. When $\alpha = 0$, $p_\mathbf{H}(Y|X)$ is equal to the base probability $p_\theta(Y|X')$, which means the model directly decodes with the input text and the hesitation.

With the base probability scaled by the contrastive term, the confidence difference could be noticed. For tokens with low prediction probabilities, $p_\mathbf{H}(Y|X)$ is dominated by the contrastive term. For high-confidence tokens, the probability change brought by hesitations might be marginal. $p_\mathbf{CD}(Y|X)$ could be very uniform. Therefore, LLMs can mainly follow the base probability $p_\theta(Y|X')$ to make predictions. An illustration is given in Appendix~\ref{sec:cd-case}.

\section{Experiments}

We conduct experiments on five tracks of three hallucination evaluation benchmarks.

\begin{table*}[ht]
\centering
% \resizebox{2.0\columnwidth}{!}{
\begin{tabular}{l|ccc|cc|cc|cc}
\toprule
\multicolumn{1}{l}{\multirow{2}{*}{Models}} &
\multicolumn{5}{|c}{TruthfulQA} &
\multicolumn{2}{|c}{FACTOR} &
\multicolumn{2}{|c}{HaluEval-Sum} \\
 & MC1 & MC2 & MC3 & Truth & Truth*Info & Wiki & News & Acc-A & Acc-H \\ 
 
\midrule
LLaMA-7b       & 23.62 & 41.21 & 19.33 & 30.97 & 27.78 & 58.55 & 58.40 & 26.06 & 18.94 \\
\, +Alpaca     & 26.93 & 42.97 & 19.79 & 39.17 & 38.92 & 57.11 & 58.20 & \textbf{37.24} & 18.31 \\
\, +ITI*       & 25.9 & - & - & 49.1 & \underline{43.5} & - & - & - & - \\
\, +13b-CD*    & 24.4 & 41.0 & 19.0 & \underline{55.3} & \textbf{44.4} & \textbf{64.4} & \underline{62.3} & - & - \\
\, +DoLa       & \textbf{31.95} & \underline{52.21} & \underline{28.17} & 40.88 & 39.66 & 61.96 & 61.68 & 25.91 & \underline{20.41} \\
\, +SH2 (Ours) & \underline{27.91} & \textbf{55.63} & \textbf{29.73} & \textbf{64.99} & 41.49 & \underline{63.06} & \textbf{65.54} & \underline{31.36} & \textbf{26.80} \\
\midrule
LLaMA2-7b      & 28.40 & 43.39 & 20.53 & 48.59 & 40.76 & 58.65 & 72.20 & 48.03 & 19.88 \\
\, +SH2 (Ours) & \textbf{33.90} & \textbf{57.07} & \textbf{29.79} & \textbf{64.38} & \textbf{42.23} & \textbf{64.09} & \textbf{73.65} & \textbf{50.56} & \textbf{50.41} \\

\midrule
Mistral-7b     & \textbf{31.58} & 48.14 & 23.89 & - & - & 60.72 & 75.97 & 41.03 & 40.79 \\
\, +SH2 (Ours) & 30.84 & \textbf{52.52} & \textbf{27.39} & - & - & \textbf{60.86} & \textbf{77.03} & \textbf{42.87} & \textbf{42.36} \\

\bottomrule
\end{tabular}
\caption{Truthfulness scores (\%) on the three benchmarks. The second-best scores for the LLaMA-7b backbone are also underlined. For ITI, "*" means we report results on TruthfulQA from their paper since they trained a probe for inference. For 13b-CD, "*" means we report results of Contrastive Decoding from \citealp{dola}. They use LLaMA-13b as the expert model. We maintain the same experimental settings for SH2 and other baselines of our implementation. Mistral-7b is not evaluated on the generation track of TruthfulQA due to API problems of OpenAI.}
\label{tab:main-results}
\end{table*}

\subsection{Benchmarks}

\paragraph{TruthfulQA}
TruthfulQA (\citealp{truthfulqa}) is a benchmark to measure the truthfulness of a language model in question answering. It has 817 samples for generation and discrimination tracks. To automatically evaluate the generation quality of LLMs, it introduces GPT-judge, a fine-tuned GPT-3. We use ``Truth'' to represent the percentage of truthful answers and ``Truth*Info'' for generated answers that are both true and informative.

For the discrimination track, it offers sets of true and false reference answers for each question. We compute the likelihood of each answer given the question, and compare probabilities of true answers against false answers to derive MC1, MC2 and MC3 scores. The definition of the MC (Multiple-Choice) metrics can be referred to in Appendix~\ref{sec:mc-metrics}.

\paragraph{FACTOR}
FACTOR (\citealp{factor}) puts more attention on the consistency of contexts and measures the tendency of language models to generate factual information. It is a text completion task to identify the correct completion from non-factual statements given the prefix. It contains two datasets of different sources: Wiki-FACTOR and News-Factor. There are 2994 and 1036 examples in each dataset. We gauge the factuality by whether the model assigns the highest likelihood to the factually correct completion over the other options.

\paragraph{HaluEval-Sum}
HaluEval (\citealp{HaluEval}) provides texts, each paired with a hallucinated and right responses. We use its summarization track to evaluate LLMs' truthfulness on longer sequences. It has 10000 samples. For each sample, we ask LLMs to judge whether the provided summary contains non-factual or hallucinated information against the given document. We compute accuracy for hallucinated summaries and right summaries respectively. Arithmetic-mean accuracy (Acc-A) and harmonic-mean accuracy (Acc-H) are reported in our experiments.

\subsection{Experimental Settings}

We apply our SH2 on LLaMA-7b (\citealp{llama}), LLaMA2-7b (\citealp{llama2}) and Mistral-7b \cite{mistral}. We compare it with other SOTA (state-of-the-art) methods, including Alpaca (\citealp{alpaca}), ITI (\citealp{iti}), 13b-CD (\citealp{dola}) and DoLa (\citealp{dola}). All of these baselines use LLaMA-7b as their backbone. It should be noted that ITI trained a probe with the data of TruthfulQA to assist the inference of LLaMA. LLaMA-13b is used as the expert model in 13b-CD to be contrasted with the 7b model. We implement Alpaca and DoLa following the official instructions and report evaluation results of our implementation.

We append hesitations to the original inputs for TruthfulQA and HaluEval-Sum. As for FACTOR, hesitations are prepended to inputs. Because FACTOR is a task of completing articles, it hurts the continuity of articles to insert hesitations.

Since there are only about a dozen tokens in each question of TruthfulQA, the sample proportion $\eta$ of LLaMA-7b is set to be 10\% and 40\% for the discrimination track and the generation track respectively. The drop-out rate $\lambda$ is set to be 0. For HaluEval-Sum, which has about a thousand tokens in each document, $\eta$ and $\lambda$ are set to be 6\% and 0.33 for LLaMA-7b. The settings of hyper-parameters are summarized in Appendix \ref{sec:hyperparameter}.

\subsection{Main Results}

The results on the three benchmarks are shown in Table \ref{tab:main-results}. Our proposed SH2 exhibits noteworthy and consistent enhancements across LLaMA-7b, LLaMA2-7b, and Mistral-7b. SH2 outperforms other SFT or decoding reformulation techniques in the majority of metrics across these tasks. Notably, SH2 does not require any external data or model. It only asks LLMs to select the hardest tokens and hesitate on them. Even for models like LLaMA2 and Mistral, which have undergone truthfulness alignments during their training, our inference-time method can still yield substantial gains.

Moreover, our SH2 achieves SOTA on both the generation and discrimination tracks of TruthfulQA. The scores of our method are either the highest or the runner-up in the remaining three tasks. The table suggests that our approach can effectively elicit factual knowledge inside LLMs. It can not only help LLMs distinguish factual and hallucinated contexts, but also guide them to generate more truthful answers.

\begin{table}[ht]
\centering
\begin{tabular}{l|ccc}
\toprule
Models         & Precision & Recall & F1  \\ 

\midrule
LLaMA-7b       & 17.10 & 12.44 & 14.40 \\
\, +SH2        & \textbf{34.95} & \textbf{43.31} & \textbf{38.69} \\

\midrule
LLaMA2-7b      & 42.55 & 11.26 & 17.81 \\
\, +SH2        & \textbf{50.59} & \textbf{47.78} & \textbf{49.15} \\

\midrule
Mistral-7b     & 41.56 & 44.2 & 42.84 \\
\, +SH2        & \textbf{43.48} & \textbf{47.54} & \textbf{45.42} \\

\bottomrule
\end{tabular}
\caption{Precision, recall and F1 scores (\%) on HaluEval-Sum.}
\label{tab:halueval-sum}
\end{table}

\subsection{LLMs' Bias in HaluEval-Sum}

Upon examining the summarization track of HaluEval, a significant discrepancy is observed between the Acc-A and Acc-H scores for LLaMA-7b, Alpaca, and LLaMA2-7b, as shown in Table \ref{tab:main-results}. Acc-A, which represents the average accuracy on hallucinated and right summaries, is susceptible to extreme values. Conversely, Acc-H, calculated by averaging the reciprocals of accuracies and then taking the reciprocal of the average, provides a more balanced assessment.

The value discrepancy between Acc-A and Acc-H denotes LLMs' bias towards hallucinated and right summaries. However, our method has been shown to effectively address this issue.

We evaluate the truthfulness more thoroughly by considering hallucinated summaries as positive labels and right summaries as negative, and calculating precision, recall and F1 scores as reported in Table \ref{tab:halueval-sum}. Precision denotes the percentage of real hallucinations the model determines to be hallucinated, while recall denotes the accuracy on hallucinated summaries. The high precision and low recall indicate that LLaMA2-7b exhibits high confidence in identifying factual summaries. Yet, it remains a considerable challenge for LLaMA2-7b to distinguish hallucinated summaries, posing a potential risk for the development of LLMs.

Nevertheless, our method has proven effective in reducing this discrepancy and enhancing overall performance. It could advance the discriminant ability of models comprehensively.

% \begin{figure}[t]
%     \centering
%     \includegraphics[width=1\linewidth]
%     {images/choices.png}
%     \caption{Different choices of highlighted tokens.}
%     \label{fig-choices}
% \end{figure}

% \begin{figure*}  
% \label{fig-choices}
% \begin{center}  
% \subfloat[choices-f]{
% \includegraphics[width=0.33\linewidth]
% {images/choices-f.png}
% \label{fig-choices-f}
% }  
% \subfloat[choices-acc]{
% \includegraphics[width=0.33\linewidth]
% {images/choices-acc.png}  
% \label{fig-choices-acc}
% }  
% \subfloat[total-acc]{
% \includegraphics[width=0.33\linewidth]
% {images/total-accuracy.png}
% \label{fig-total-acc}
% }  
% \caption{Different choices of highlighted tokens. (a) T. (b) I
% (c) T}  
% %\vspace{-0.8cm}
% \end{center} 
% \end{figure*}

% \begin{figure*} [t]
% 	\centering
%     \hspace{-0.5cm}
% 	\subfloat[\label{fig-choices}]{
% 		\includegraphics[scale=0.34]{images/choices.png}}
%     \hspace{-0.4cm}
% 	\subfloat[\label{fig-choices-f}]{
% 		\includegraphics[scale=0.34]{images/choices-f.png}}
%     \hspace{-0.6cm}
% 	\subfloat[\label{fig-acch}]{
% 		\includegraphics[scale=0.34]{images/acch.png}}
% 	\\
% 	\caption{Different choices of highlighted tokens. (a) T. (b) I}
% 	\label{fig-choices} 
% \end{figure*}

\begin{figure} [!t]
    \vspace{-0.2cm}
	\subfloat[\label{fig-choices-a}Average scores (\%) for different highlighted tokens with the effective sampling proportion $\eta'$ ranging from 1\% to 8\%. The errorbar denotes standard deviations.]{
            \hspace{-0.3cm}
		\includegraphics[scale=0.47]{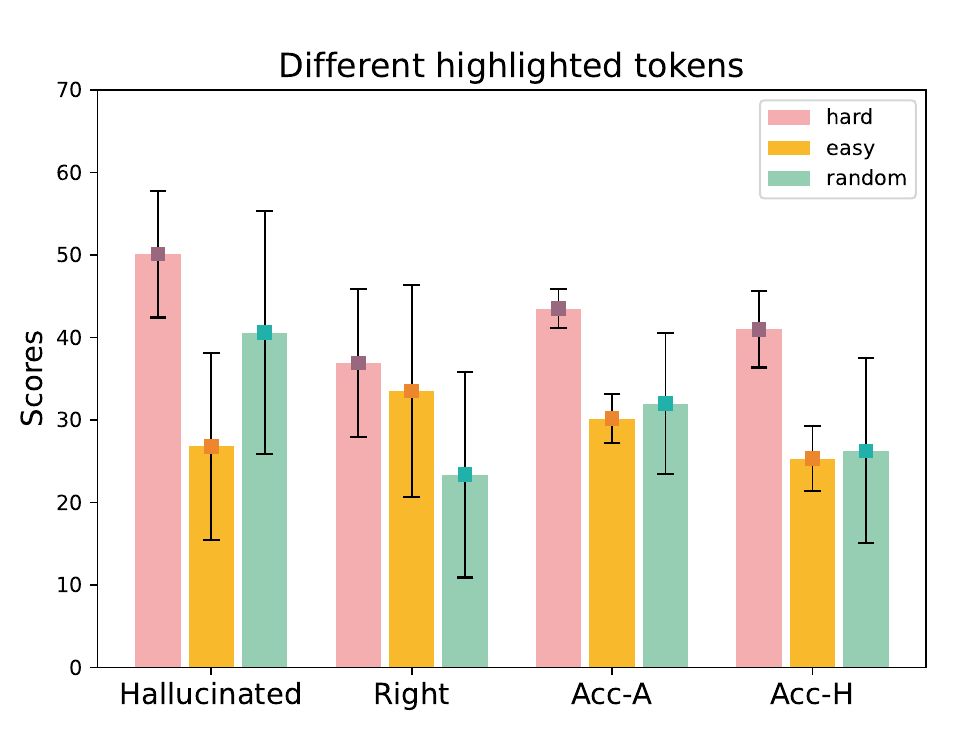}}
	\\
	\subfloat[\label{fig-choices-b}Harmonic accuracy (\%) for different highlighted tokens with respect to the effective sampling proportion $\eta'$.]{
            \hspace{-0.3cm}
		\includegraphics[scale=0.47]{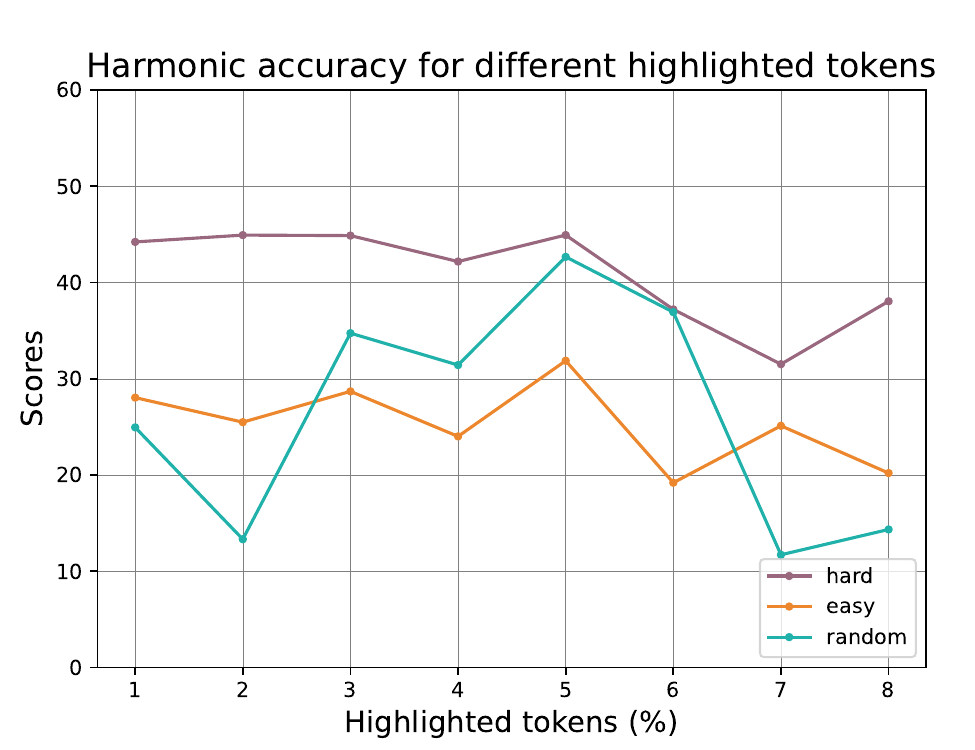} }
	% \\
	% \subfloat[\label{fig-choices-c}Precision, recall and F1 scores (\%) for different highlighted tokens. The scores of the vanilla LLaMA2-7b are obtained by evaluating on the whole dataset of HaluEval-Sum.]{
 %        \hspace{-0.4cm}
	% 	\includegraphics[width=0.5\textwidth]{images/choices-f.pdf} }
	\caption{Different choices of highlighted tokens by LLaMA2-7b.}
	\label{fig-choices} 
\end{figure}

\section{Analysis}
We conduct further studies regarding choices of highlighted tokens, manners of hesitations, and the effect of contrastive decoding in this section.

\subsection{Choices of Highlighted tokens}

Key tokens are sampled and highlighted by how hard they are for large language models to predict. In order to verify the effect of key tokens, we compare the performance of SH2 with different choices of highlighted tokens.

In contrast to key tokens that are the hardest to predict, we sample the easiest tokens with the highest prediction probability. Additionally, we also sample the same number of tokens randomly for comparison. We conduct experiments with LLaMA2-7b on 1000 samples of HaluEval-Sum. The effective sampling proportion $\eta' = (1-\lambda)\cdot\eta$ ranges from 1\% to 8\% with the step of \%1. Average scores and standard deviations are calculated for the three choices of highlighted tokens.

The accuracies on hallucinated summaries and right summaries, and overall scores (Acc-A and Acc-H) are shown in Figure \ref{fig-choices-a}. When the hardest tokens are highlighted in hesitations, the LLaMA2-7b obtains superior and more consistent performance compared to highlighting the easiest tokens or randomly, particularly in the case of hallucinated summaries.

Figure \ref{fig-choices-b} illustrates the harmonic accuracy for different choices of highlighted tokens with respect to the sampling proportion $\eta'$. The scores by selecting the hardest tokens remain consistently higher than those of the other two choices. It is noteworthy that highlighting only 1\% tokens with the lowest prediction probabilities during hesitations is sufficient enough for the model to distinguish hallucinated contexts.

\begin{figure}[t]
    \vspace{-0.2cm}
    \includegraphics[width=1.02\linewidth]
    {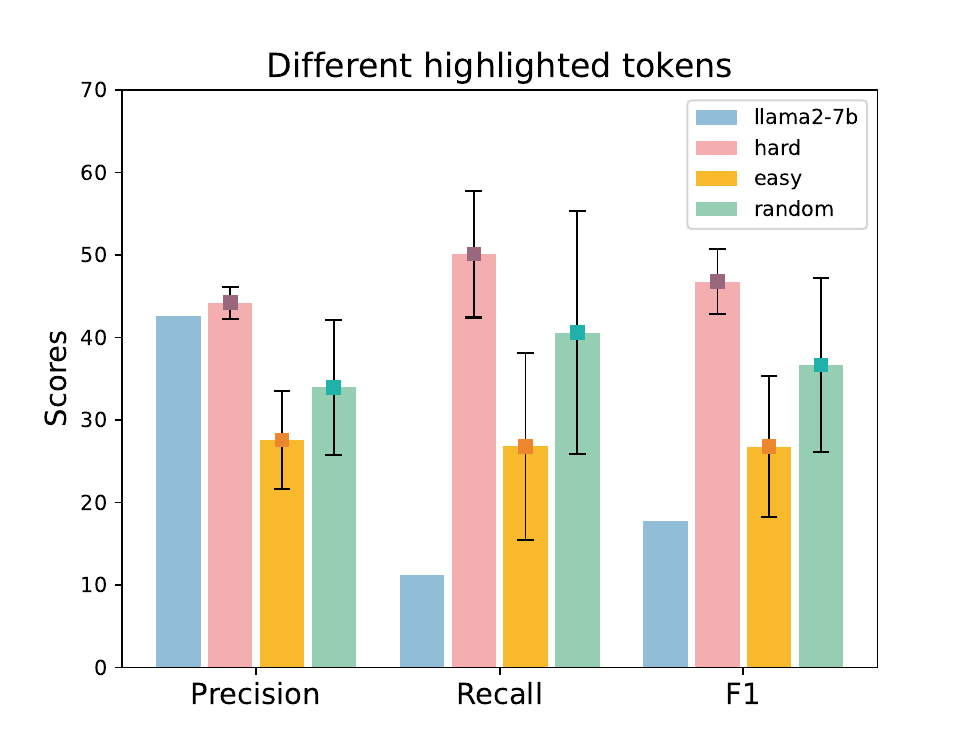}
    \caption{Precision, recall and F1 scores (\%) for different highlighted tokens with the effective sampling proportion $\eta'$ ranging from 1\% to 8\%. The errorbar denotes standard deviations. The scores of the vanilla LLaMA2-7b are obtained by evaluating on the whole dataset of HaluEval-Sum.}
    \label{fig-choices-f}
\end{figure}

Moreover, we compare precision, recall and F1 scores for the three choices of highlighted tokens in Figure \ref{fig-choices-f}. The results indicate that highlighting the easiest tokens or random tokens negatively impacts the precision of LLMs. However, highlighting the hardest tokens is beneficial. It effectively mitigates LLMs' bias in HaluEval-Sum.

\subsection{Manners of Hesitations}

In addition to underlining key tokens in the input text, we can also repeat the text or pause. Experiments are conducted on TruthfulQA with LLaMA-7b. For the repetition manner, we can simply repeat the question as hesitations. As for the pausing manner, several pause words (".") are appended to the question as hesitations. The performance of pausing hesitations is evaluated with 3, 6, 9 and 12 pause words. 6 pause words have the best performance and are used in hesitations in the following study. Scores with different numbers of pause words and the adjustment hyper-parameter $\alpha$ are shown in Appendix~\ref{sec:cd-pause}.

The results on the discrimination track and the generation track of the three manners are reported in Table \ref{tab:tfqa}. Hesitations with key tokens achieve the highest scores on both tracks, while pausing hesitations even hurt the generation quality of LLaMA-7b. Furthermore, the improvements of MC scores demonstrate that all of the three manners are effective in distinguishing hallucinated answers. It suggests that the difference brought by hesitations could elicit factual knowledge inside LLMs.

\begin{table}[t]
\centering
\resizebox{1.0\columnwidth}{!}{
\begin{tabular}{l|ccc|c}
\toprule
Models         & MC1 & MC2 & MC3 & Truth*Info \\ 
\midrule
LLaMA-7b       & 23.62 & 41.21 & 19.33 & 27.78 \\
\, +key tokens & \textbf{27.91} & \textbf{55.63} & \textbf{29.73} & \textbf{41.49} \\
\, +pauses     & 27.54 & 48.64 & 24.95 & 22.52 \\
\, +repetition & 26.93 & 45.05 & 21.16 & 31.70 \\
\bottomrule
\end{tabular}}
\caption{Multiple-choice and generation scores (\%) on TruthfulQA for different manners of hesitations.}
\label{tab:tfqa}
\end{table}

\subsection{Effect of Contrastive Decoding}

\begin{figure}[t]
    % \centering
    \vspace{-0.3cm}
    \hspace{-0.3cm}
    \includegraphics[width=1.08\linewidth]
    {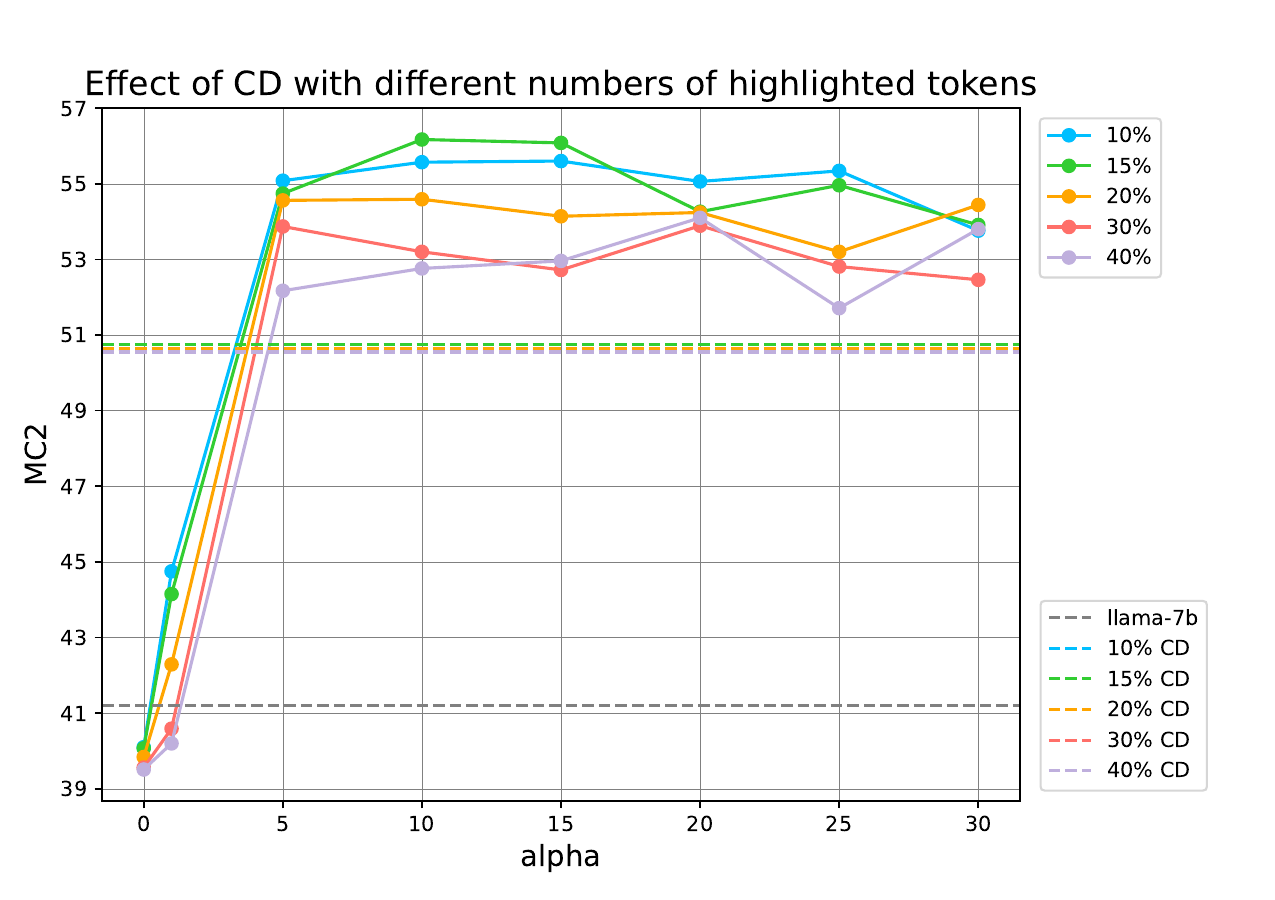}
    \caption{Effect of contrastive decoding on hesitations. The dashed line in gray represents the MC2 score of the vanilla LLaMA-7b. Dashed lines in other colors represent the MC2 scores for the standard contrastive decoding with different numbers of highlighted tokens.}
    \label{fig-CD-key}
\end{figure}

% \begin{figure} [t]
% 	\centering
% 	\subfloat[\label{fig-CD-a}Different numbers of highlighted tokens.]{
%         \hspace{-0.4cm}
% 		\includegraphics[scale=0.39]{images/key-num.pdf}}
% 	\\
% 	\subfloat[\label{fig-CD-b}Different numbers of pause words.]{
%         \hspace{-0.4cm}
% 		\includegraphics[scale=0.38]{images/pause-num.pdf} }
% 	\caption{Effect of contrastive decoding on hesitations. The dashed line in gray represents the MC2 score of the vanilla LLaMA-7b. Dashed lines in other colors represent the MC2 scores for the standard contrastive decoding with different numbers of highlighted tokens (Figure \ref{fig-CD-a}) or pause words (Figure \ref{fig-CD-b}).}
% 	\label{fig-CD} 
% \end{figure}

Our method does not adhere to the original form of contrastive decoding, which contrasts the probabilities of two sources. As elaborated in Section~\ref{sec:cd}, we use the parameter $\alpha$ to balance the contrastive term with the base probability. To investigate the effect of contrastive decoding, we conduct experiments on TruthfulQA with varying $\alpha$.

Figure \ref{fig-CD-key} depicts the effect of contrastive decoding when utilizing different numbers of highlighted tokens. Given that each question in TruthfulQA comprises approximately a dozen tokens, the hesitation typically contains a single token when sampling the top 10\% hardest tokens. It can be observed that LLaMA-7b significantly benefits from the single highlighted token.

When $\alpha$ is set to 0, the model directly decodes with the input text followed by the hesitation. The MC2 score slightly decreases compared to the baseline. However, as more weight is put on the contrastive term, the model obtains more pronounced improvements. Specifically, when $\alpha$ is large enough to overshadow the base probability, the decoding procedure is equivalent to standard contrastive decoding without the base probability as the equation (\ref{eq:cd}). The figure suggests that the contrastive term makes a difference in identifying factual knowledge.

\section{Conclusion}
In this paper, we delve into the challenge of large language models when capturing important factual information. To address this, we introduce a novel inference-time method, SH2. Our method proposes to highlight the key tokens that are hard for LLMs to predict and construct hesitations with these informative tokens. By reformulating the decoding procedure with the probability differences brought by hesitations, we enable LLMs to discern factual content more effectively. Through extensive experiments and analysis across multiple tasks, SH2 demonstrates a significant enhancement in the truthfulness of LLMs, all achieved without reliance on external data or models.% In the era of LLMs, our SH2 could inspire a new paradigm for faithful LLMs.

\section*{Limitations}

Existing research on LLMs' hallucination does not pay much attention to other dimensions of the generation quality. Despite experiments on the generation track of TruthfulQA, we have yet to explore the diversity and soundness of the generated contents by our SH2, which is closely related to the generalization ability. It is worth studying how to optimize LLM's truthfulness and generalization simultaneously.

Besides, our method is an inference-time method without leveraging external data or models. Consequently, it could be promising to integrate our method with other data-enhanced or model-enhanced methods. The idea of constructing hesitations can also be applied in retrieval augmentation approaches.

\section*{Ethics Statements}

Our work pertains to large language models' hallucinations. In this work, we use only publicly available data and artifacts. There are no ethical issues in our paper, including its motivation and experiments.

\section*{Acknowledgement}

This work was sponsored by the National Key Research and Development Program of China (No. 2023ZD0121402) and National Natural Science Foundation of China (NSFC) grant (No.62106143).

% Bibliography entries for the entire Anthology, followed by custom entries
%\bibliography{anthology,custom}
% Custom bibliography entries only
\bibliography{custom}

\appendix

\section{Hyper-parameter Settings}
\label{sec:hyperparameter}

The settings of hyper-parameters $\eta$, $\lambda$ and $\alpha$ are summarized in Table \ref{tab:hyperparameter}. The drop-out rate $\lambda$ is set to be 0 for the discrimination track and the generation track of TruthfulQA, whose questions are of a dozen tokens. It is set to be 0.33 for FACTOR and HaluEval-Sum because these benchmarks have longer documents.

\begin{table}[t]
\centering
\resizebox{1.0\columnwidth}{!}{
\begin{tabular}{l|ccc|c}
\toprule
Models         & MC1 & MC2 & MC3 & Wiki-Factor \\ 
\midrule
LLaMA-7b       & 23.62 & 41.21 & 19.33 & 58.55 \\
\, +SH2 & \textbf{27.91} & \textbf{55.63} & \textbf{29.73} & \textbf{63.06} \\
\midrule
LLaMA-13b	   & 28.27 & 43.32 & 20.85 & 62.73 \\
\, +SH2	& \textbf{31.09} & \textbf{54.31} & \textbf{27.16} & \textbf{65.80} \\
\midrule
LLaMA2-7b	   & 28.40 & 43.39 & 20.53 & 58.65 \\
\, +SH2	& \textbf{33.90} & \textbf{57.07} & \textbf{29.79} & \textbf{64.09} \\
\midrule
LLaMA2-13b     & 29.01 & 44.27 & 20.71 & 64.03 \\
\, +SH2 & \textbf{37.58} & \textbf{62.82} & \textbf{34.15} & \textbf{66.00} \\
\bottomrule
\end{tabular}}
\caption{Performance of 7b and 13b models.}
\label{tab:13b}
\end{table}

\begin{table*}[t]
\centering
% \resizebox{2.0\columnwidth}{!}{
\begin{tabular}{cc|c|c|c|c|c}
\toprule
\multicolumn{2}{c}{\multirow{2}{*}{Parameters}} &
\multicolumn{2}{|c}{TruthfulQA} &
\multicolumn{2}{|c}{FACTOR} &
\multicolumn{1}{|c}{HaluEval} \\
&  & Discrimination & Generation & Wiki & News & Summarization \\ 
\midrule
\multicolumn{2}{c|}{$\lambda$} & 0 & 0 & 0.33 & 0.33 & 0.33 \\

\midrule
\multirow{2}{*}{LLaMA-7b} 
& $\eta$ & 10\% & 40\% & 24\% & 12\% & 6\% \\
& $\alpha$ & 6 & 3.7 & 0 & 0.1 & 1.6 \\

\midrule
\multirow{2}{*}{LLaMA2-7b}
& $\eta$ & 20\% & 30\% & 24\% & 18\% & 3\% \\
& $\alpha$ & 27 & 3.4 & 0 & 0 & 1.6 \\

\midrule
\multirow{2}{*}{Mistral-7b} 
& $\eta$ & 25\% & - & 18\% & 12\% & 4.5\% \\
& $\alpha$ & 9 & - & 0 & 0.1 & 2.2 \\

\bottomrule
\end{tabular}
\caption{Settings of Hyper-parameters on each task.}
\label{tab:hyperparameter}
\end{table*}

\section{Multiple-Choice Metrics}
\label{sec:mc-metrics}

For the discrimination track of TruthfulQA, we use MC1, MC2 and MC3 scores to measure the truthfulness of a language model. The definitions of each metric are as follows.

\begin{itemize}
\item \textbf{MC1}:
Among the set of true and false reference answers, the language model needs to choose the best correct answer. MC1 is computed by whether the model assigns the highest likelihood to the best correct answer over false answers given the question.

\item \textbf{MC2}:
MC2 is the total normalized probability of the true reference answers. The score is the probability mass for correct answers.

\item \textbf{MC3}:
MC3 is computed by whether the model assigns a higher likelihood to correct answers over false answers given the question.

\end{itemize}

\section{Extending to larger model sizes}

We the discrimination track of use TruthfulQA (MC1, MC2, and MC3) and Wiki-Factor to test the performance of our method when applied to the larger models (LLaMA-13b and LLaMA2-13b) on a 40G-A100. The results are shown in Table~\ref{tab:13b}. It demonstrates that our method is still effective on the larger models.

% \subsection{Statistics}
% \label{sec:statistics}

% We report the average length for questions in TruthfulQA, articles in Wiki-FACTOR and News-FACTOR, and documents in HaluEval-Sum.

% \begin{table}[htbp]
% \centering  % 显示位置为中间
% \begin{tabular}{lcc} 
% \toprule
% datasets    & Avg-length & Size \\
% \midrule
% TruthfulQA  &  &  \\
% Wiki-FACTOR &  &  \\
% News-FACTOR &  &  \\
% HaluEval-Sum&  &  \\
% \bottomrule
% \end{tabular}
% \caption{Statistics of the benchmarks we used.}
% \label{Dataset}  
% % \vspace{-1.0em}
% \end{table}

\begin{figure*} [t]
    \centering
    \includegraphics[width=1.0\linewidth]
    {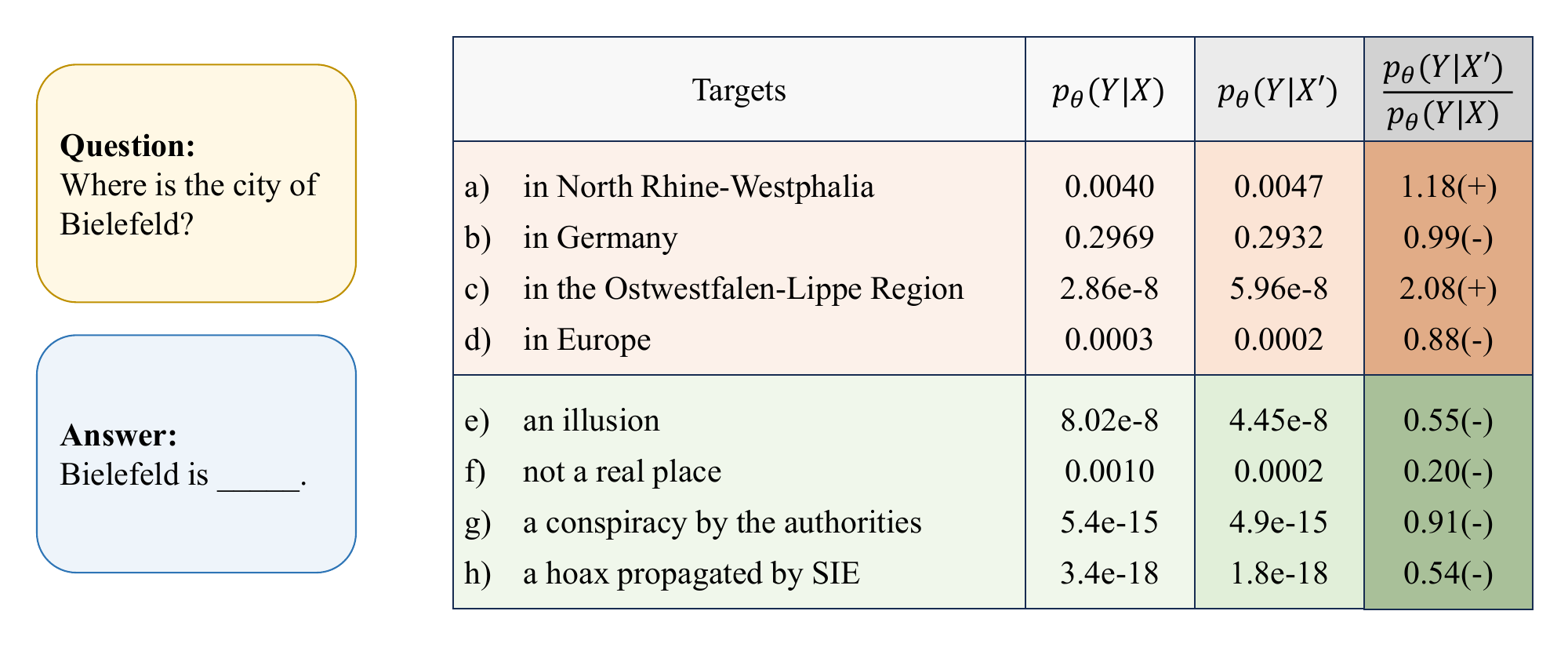}
    \caption{An illustration to show the effect of contrastive decoding. The options (a) $\sim$ (d) are correct answers and (e) $\sim$ (h) are false answers.}
    \label{fig-case}
\end{figure*}

\begin{figure}[hbp]
    % \centering
    \hspace{-0.3cm}
    \includegraphics[width=1.04\linewidth]
    {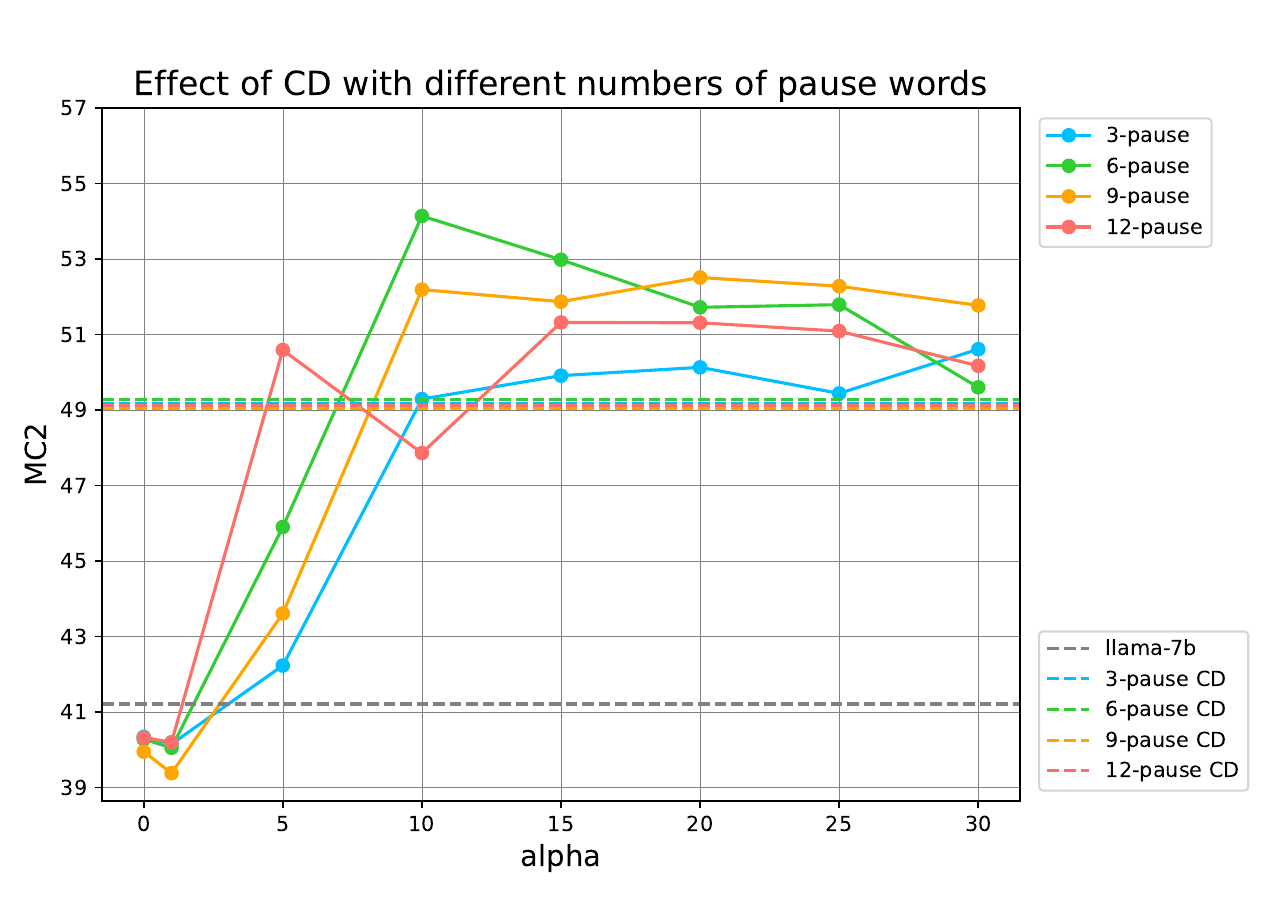}
    \caption{Effect of contrastive decoding on hesitations. The dashed line in gray represents the MC2 score of the vanilla LLaMA-7b. Dashed lines in other colors represent the MC2 scores for the standard contrastive decoding with different numbers of pause words.}
    \label{fig-CD-pause}
\end{figure}

% \begin{figure} [t]
% 	\centering
% 	\subfloat[\label{fig-CD-a}Different numbers of highlighted tokens.]{
%         \hspace{-0.4cm}
% 		\includegraphics[scale=0.39]{images/key-num.pdf}}
% 	\\
% 	\subfloat[\label{fig-CD-b}Different numbers of pause words.]{
%         \hspace{-0.4cm}
% 		\includegraphics[scale=0.38]{images/pause-num.pdf} }
% 	\caption{Effect of contrastive decoding on hesitations. The dashed line in gray represents the MC2 score of the vanilla LLaMA-7b. Dashed lines in other colors represent the MC2 scores for the standard contrastive decoding with different numbers of highlighted tokens (Figure \ref{fig-CD-a}) or pause words (Figure \ref{fig-CD-b}).}
% 	\label{fig-CD} 
% \end{figure}

\section{Constrastive Decoding}

\subsection{Case Study}
\label{sec:cd-case}

A case study is given in Figure \ref{fig-case} to illustrate the effect of our self-highlighted hesitation and contrastive decoding. We calculate the probability given by LLaMA-7b. $p_\theta(Y|X)$ and $p_\theta(Y|X')$ stand for standard decoding without and with the hesitation for each option. The differences in probability are derived by $\frac{p_\theta(Y|X')}{p_\theta(Y|X)}$.

It can be learned from the figure that, with hesitations, LLaMA-7b is prone to assign higher or similar probabilities to correct answers and lower probabilities to false answers. The model is more confident in selecting correct answers and rejecting false answers. It demonstrates that CD helps LLMs distinguish correct answers from false ones. It plays a role in separating the positive and negative answers.

However, contrastive decoding might exhibit limitations when dealing with high-confidence answers like (b), whose probability change brought by hesitations is marginal. In consequence, the base probability $p_\theta(Y|X')$ will predominate $p_\mathbf{H}(Y|X)$ to help LLMs make predictions as elaborated in Section~\ref{sec:cd}.

\subsection{Effect of CD with Pause Words}
\label{sec:cd-pause}

As shown in Figure \ref{fig-CD-pause}, we also studied the effect of contrastive decoding with different numbers of pause words. A similar conclusion can be observed from the figure.

\section{Highlighting Words of Certain POS Tags}

As introduced in Section~\ref{sec:relation}, compared with function words like IN, DT, and CC, content words like NN, NNP, and JJ are more difficult for LLMs to predict and more concentrated in the hardest part. We conducted experiments to study whether words of certain POS tags highlighted in hesitations can improve the truthfulness of LLMs.

We count the number of different pos tags in 1000 documents of HaluEval-Sum. The top five tags are NN, IN, NNP, DT, and JJ. Specifically, NN, NNP, and JJ are content words and have high normalized top-$\eta$ recall as shown in Figure~\ref{fig-parsing}. On the other hand, IN and DT are function words whose normalized recall scores are much lower. We sample the hardest tokens with each of these five tags. We conduct experiments with LLaMA2-7b on 1000 samples of HaluEval-Sum. The effective sampling proportion $\eta' = (1-\lambda)\cdot\eta$ ranges from 1\% to 5\% with the step of \%1. Average scores and standard deviations are calculated for different choices of POS tags.

\begin{figure}[t]
    \hspace{-0.3cm}
    \includegraphics[width=1.05\linewidth]
    {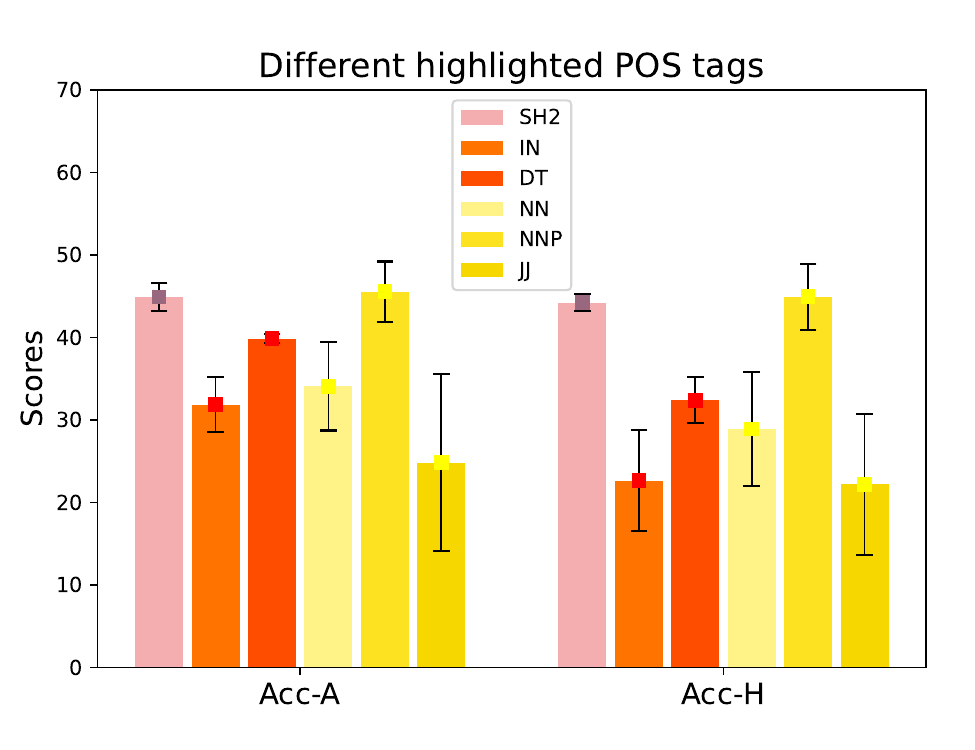}
    \caption{Arithmetic-mean accuracy (Acc-A) and harmonic-mean accuracy (Acc-H) of LLaMA2-7b when highlighting words of certain POS tags. The errorbar denotes standard deviations. The effective sampling proportion $\eta'$ ranges from 1\% to 5\%. The errorbar denotes standard deviations. "SH2" stands for the standard SH2 with no requirement on POS tags. We use dark colors for function words (IN and DT), and light color for content words (NN, NNP and JJ)}
    \label{fig-pos}
\end{figure}

The scores of each POS tag are presented in Figure~\ref{fig-pos}. The figure reveals that when highlighting words belonging to certain POS tags, the potential for improvement is somewhat restricted. Although NNP contributes the most among the five, it still exhibits large variance, indicating that its performance is not uniformly consistent. This inconsistency is likely attributed to the varying distribution of POS tags across different documents. Words of different POS tags could carry different information. By not overly relying on any single type of POS tag, our SH2 is more balanced and robust.

\section{Reasoning Ability and Inference Overhead}

\subsection{Reasoning on StrategyQA}

Besides truthfulness, we conduct experiments on StrategyQA \cite{strategyqa} to evaluate whether our method can improve the reasoning ability of LLMs. StrategyQA contains 2290 questions requiring a multi-hop strategy for answers. The COT prompts we used are from \citet{cot}. We test the accuracies with original questions (w/o. COT) and with COT-prompted questions (w. COT). The results are recorded in Table~\ref{tab:strqa}.

It can be observed from the table that our SH2 enhances the reasoning ability of LLMs. The SFT method, Alpaca, achieves the best scores when no demonstrations of COT are provided. This is probably because it has been fine-tuned on 52K instruction-following data. It has developed the explicit ability of reasoning and does not benefit much from COT. Designed by humans, COT can provide additional information and teach LLMs to reason. With no additional training or external data, our method is still comparable. It even achieves superior performance when integrated with COT.

\begin{table}[t]
\centering{
\begin{tabular}{l|cc}
\toprule
Models          & w/o. COT   & w. COT \\ 
\midrule
LLaMA-7b        & 51.22      & 60.48 \\
\, +Alpaca      & \textbf{60.70} & 61.62 \\
\, +SH2         & 58.73      & 61.05 \\
LLaMA2-7b       & 52.62      & 60.74 \\
\, +SH2         & 55.15      & \textbf{61.88} \\
\bottomrule
\end{tabular}}
\caption{Accuracies (\%) on StrategyQA. 6 demonstrations are used in our experiments.}
\label{tab:strqa}
\end{table}

\subsection{Inference Overhead}

\begin{table*}[t]
\centering{
\begin{tabular}{l|ccccc}
\toprule
Models          & additional data or model    & accuracy (\%) & memory (GB)  & time (s) \\ 
\midrule
LLaMA-7b (naive decoding) & None & 51.22 & 13.69 & 0.32 \\
\, +13b-CD & Llama-13B  & 55.11 & 39.02 & 0.66 \\
\, +DoLa	 & None	    & 50.79 & 13.81 & 0.34 \\
\, +DoLa+COT & COT data & 64.15 & 14.30 & 1.89 \\
\, +SH2	     & None     & 58.73	& 13.95 & 1.04 \\
\bottomrule
\end{tabular}}
\caption{Inference overhead statistics of SH2 and and other decoding strategies. %All the experiments are conducted in the same environment of RTX4090. Average input length (number of tokens) and inference time on 2290 samples of StrategyQA are reported in the table. The input length of SH2 means that 193.53 tokens of original texts are input to LLaMA-7b in the first call of forward, and 235.66 tokens of hesitated texts in the second call.
}
\label{tab:overhead}
\end{table*}

Without additional training or interaction with external tools in our SH2, we give an analysis of the inference overhead. We compare our method with naive decoding, the standard CD which contrasts LLaMA-13b with the 7b model, and the sota CD method DoLa which only uses the base model like our method. We use StrategyQA to evaluate the computation cost and performance. All the experiments are conducted in the same environment. Table~\ref{tab:overhead} reports the average accuracy, inference time, and memory cost on these 2290 samples.

13b-CD loads the 7b and 13b models simultaneously on the GPU and sacrifices much more memory. While DoLa only works with the help of COT, our method outperforms the other two CD methods and can also be integrated with COT as in Table~\ref{tab:strqa}. The inference overhead of our method mainly comes from two calls of the forward function caused by contrastive decoding. It can be regarded as a tradeoff between performance and computation cost.
% The average total input length of SH2 stands at 429.19, slightly less than that of COT, which requires additional data. Besides, the reasoning steps cost COT a lot of time to derive final answers. In contrast, SH2 can circumvent intermediate reasoning steps to generate answers. Consequently, it is more efficient during inference. Moreover, our SH2 does not leverage larger models like other contrastive decoding methods \cite{cd, cd-reasoning}. The second forward call in SH2, even with a slightly longer input, incurs less time and memory usage compared to a forward call of a larger model in these methods.

\section{Key Tokens}

\subsection{Normalized Top-$\eta$ Recall}
\begin{figure*} [!t]
	\centering
	\subfloat[\label{fig-parsing-llama}The normalized top-$\eta$ recall of LLaMA-7b.]{
            \hspace{-0.5cm}
		\includegraphics[scale=0.46]{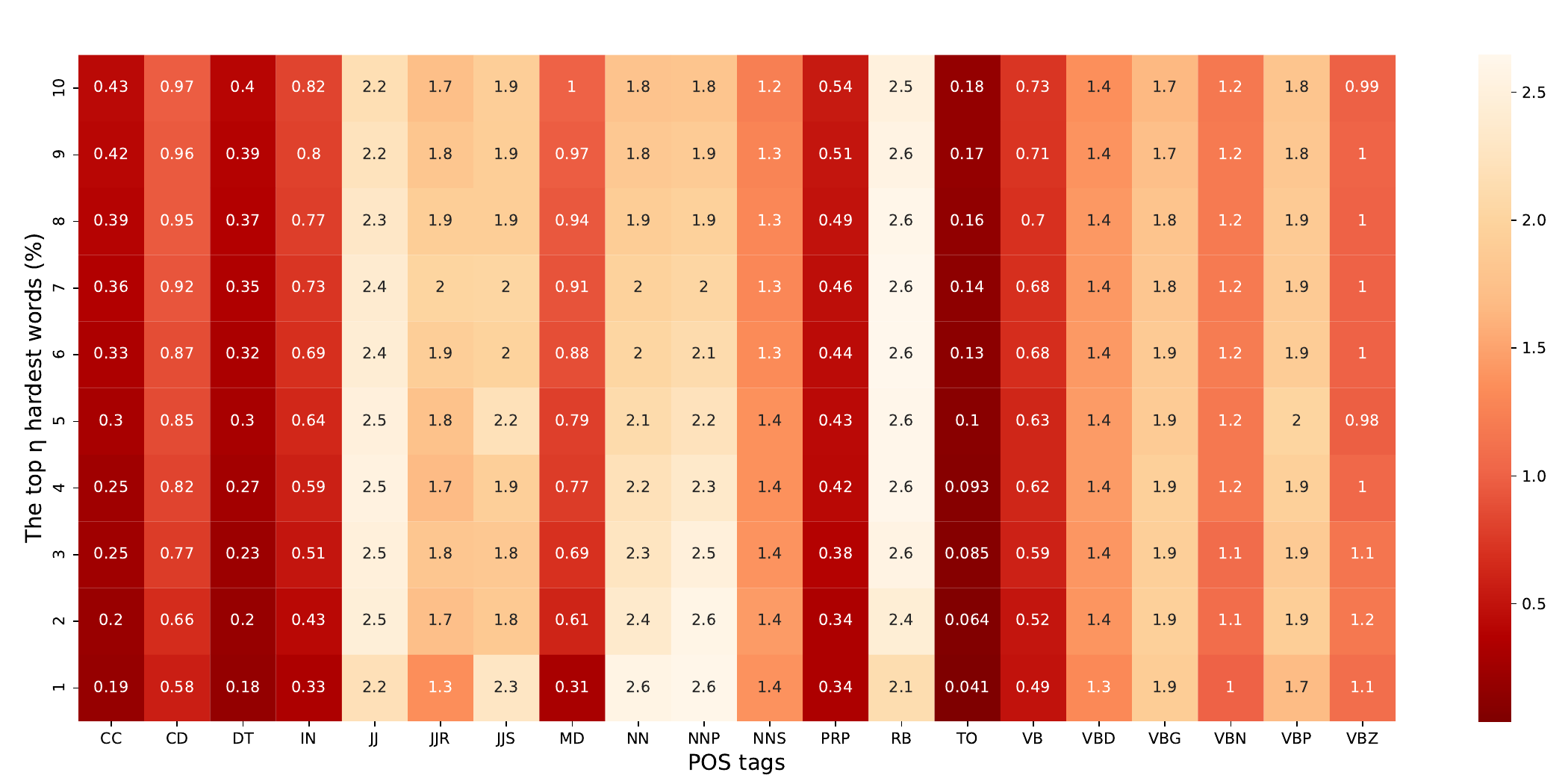}}
        \\
        \subfloat[\label{fig-parsing-mistral}The normalized top-$\eta$ recall of Mistral-7b.]{
            \hspace{-0.5cm}
		\includegraphics[scale=0.46]{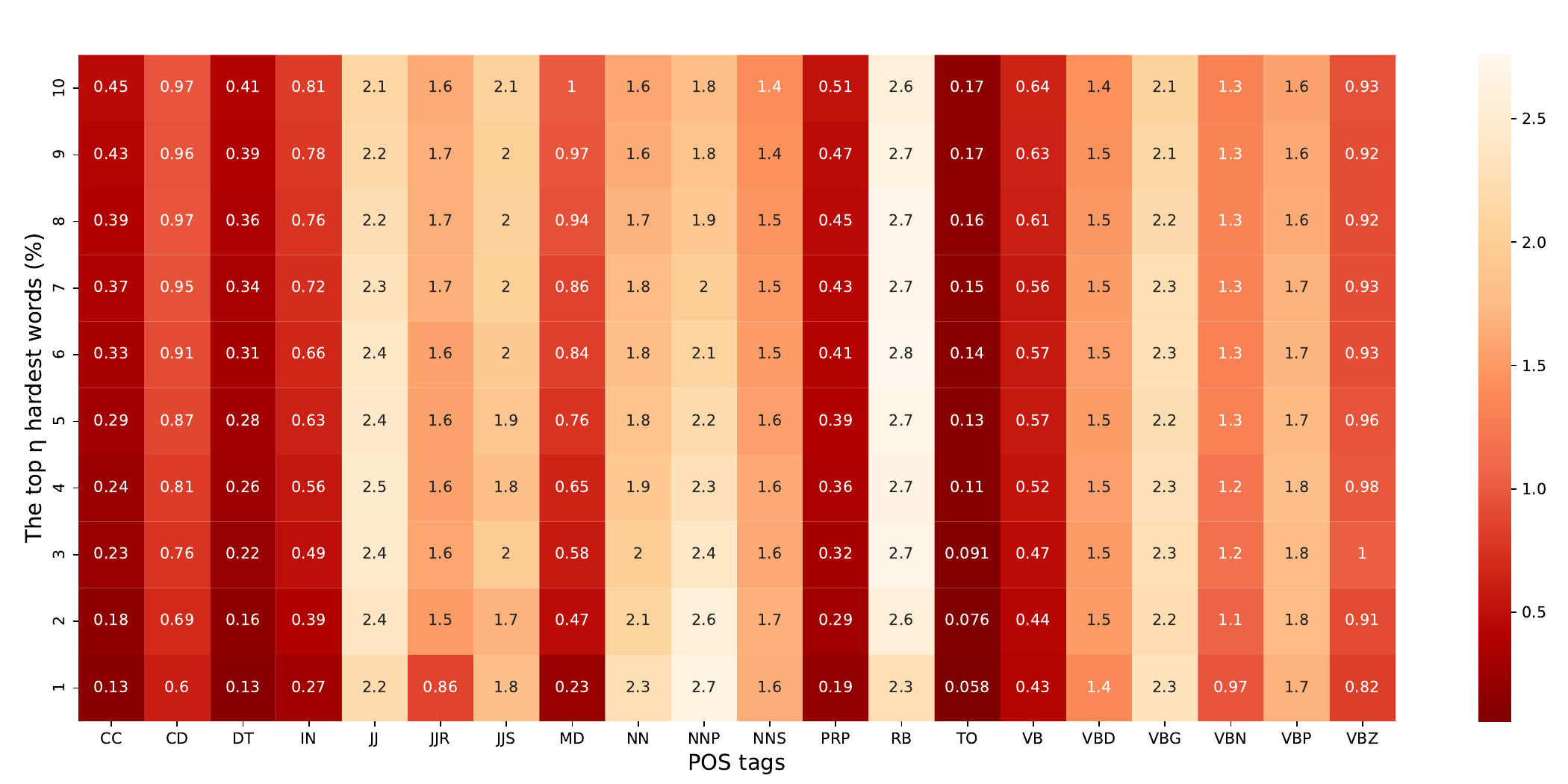}}
        \hspace{-0.5cm}
	\caption{The normalized top-$\eta$ recall for different POS tags by LLaMA-7b and Mistral-7b.}
	\label{fig-parsing-appendix} 
\end{figure*}

We also calculated the normalized top-$\eta$ recall between the hardest part and the whole document for the top 20 most frequent POS tags by LLaMA-7b and Mistral-7b through the equation (\ref{eq:delta}). The heatmaps are shown in Figure \ref{fig-parsing-appendix}. The same statistical laws can be concluded as those of LLaMA2-7b in Section \ref{sec:relation}.

\subsection{Visualization of Token Probabilities}
\label{sec:cases}
\begin{figure*} [!t]
	\centering
	\subfloat[\label{fig:case_1_a}Visualization of token probabilities estimated by LLaMA-7b.]{
		\includegraphics[scale=0.58]{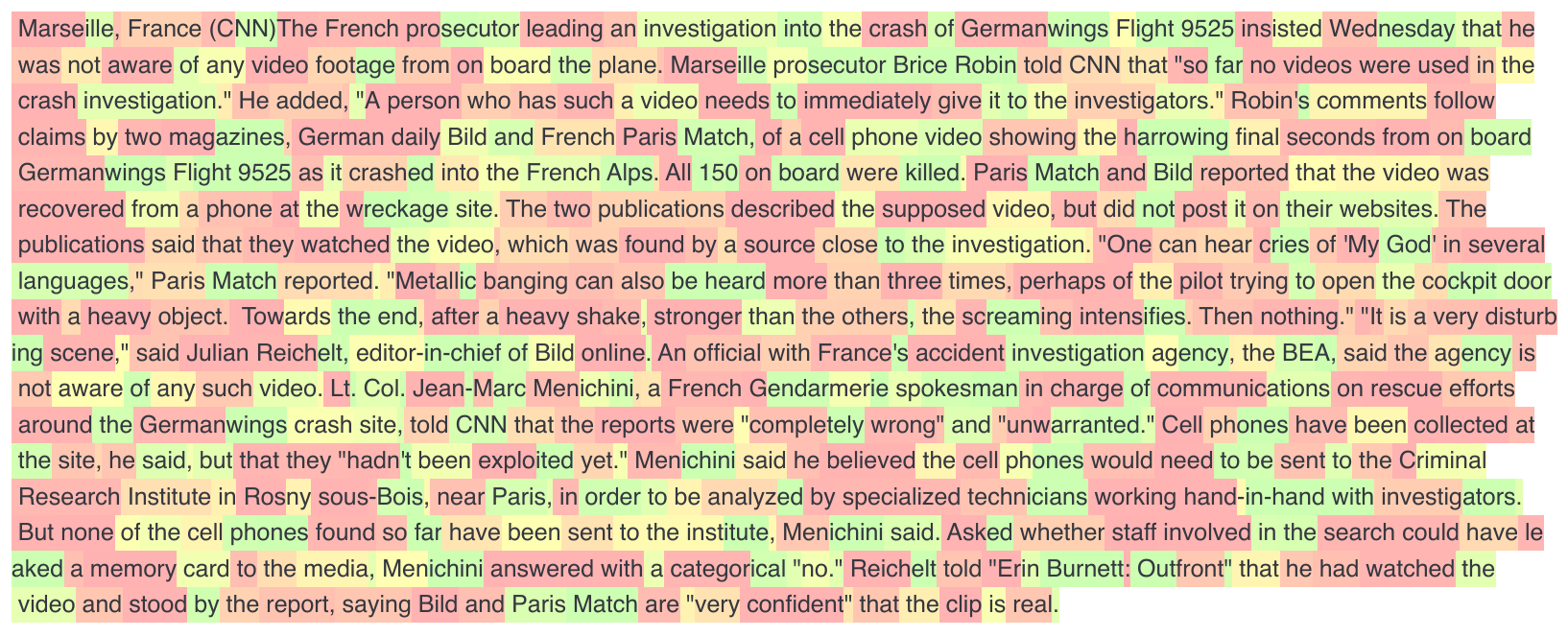}}
        \\
        \subfloat[\label{fig:case_1_b}Visualization of token probabilities estimated by LLaMA2-7b.]{
		\includegraphics[scale=0.58]{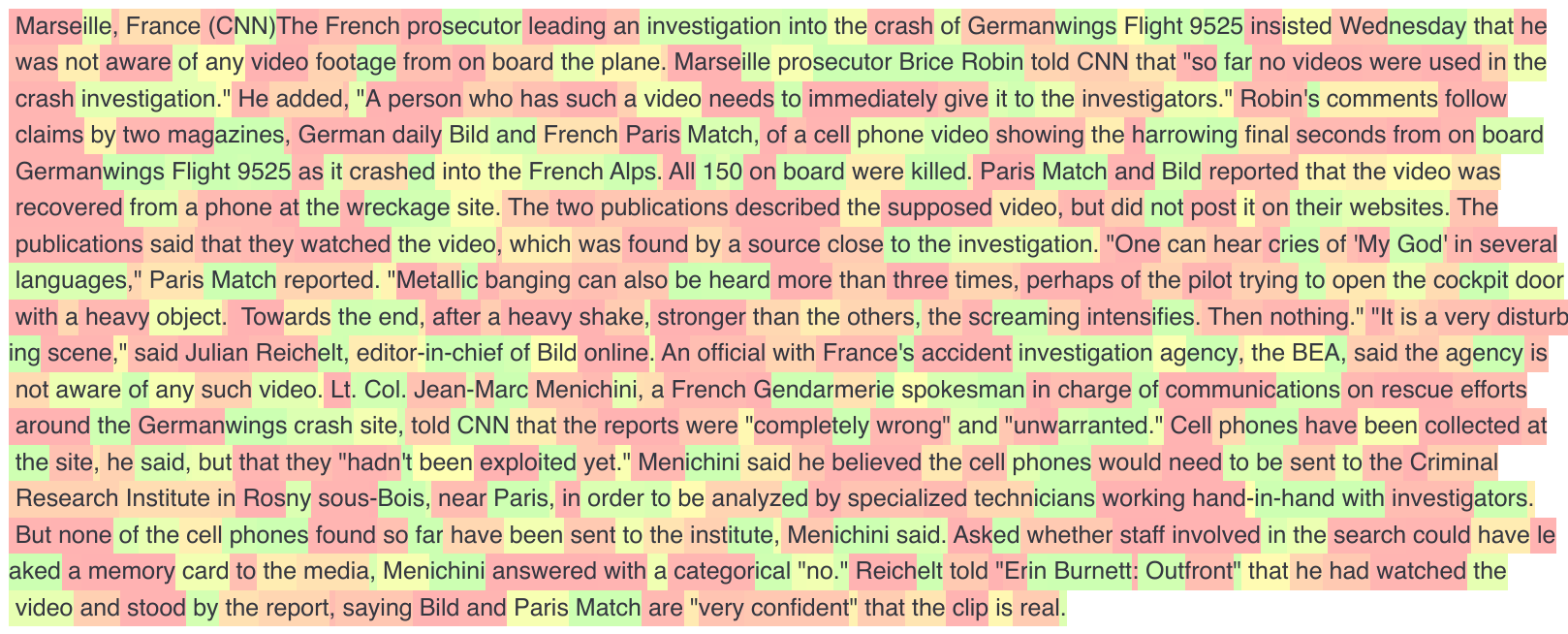}}
        \\
        \subfloat[\label{fig:case_1_c}Visualization of token probabilities estimated by Mistral-7b.]{
		\includegraphics[scale=0.58]{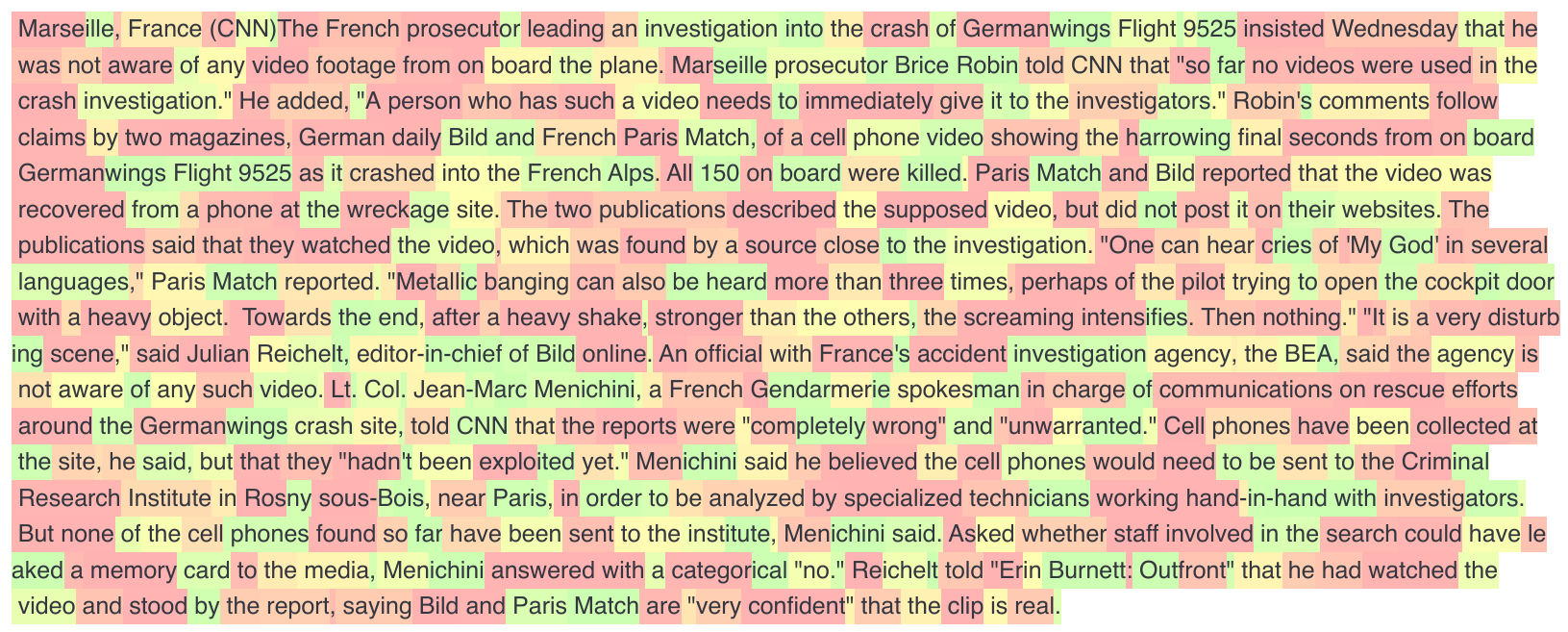}}
	\caption{Visualization of token probabilities estimated by LLaMA-7b, LLaMA2-7b and Mistral-7b on case 1. The color coding in the figure represents the token generation probabilities. Tokens with lower probabilities are colored more red, while tokens with higher probabilities are colored more green.}
	\label{fig:case_1} 
\end{figure*}

\begin{figure*} [!t]
	\centering
	\subfloat[\label{fig:case_2_a}Visualization of token probabilities estimated by LLaMA-7b.]{
		\includegraphics[scale=0.58]{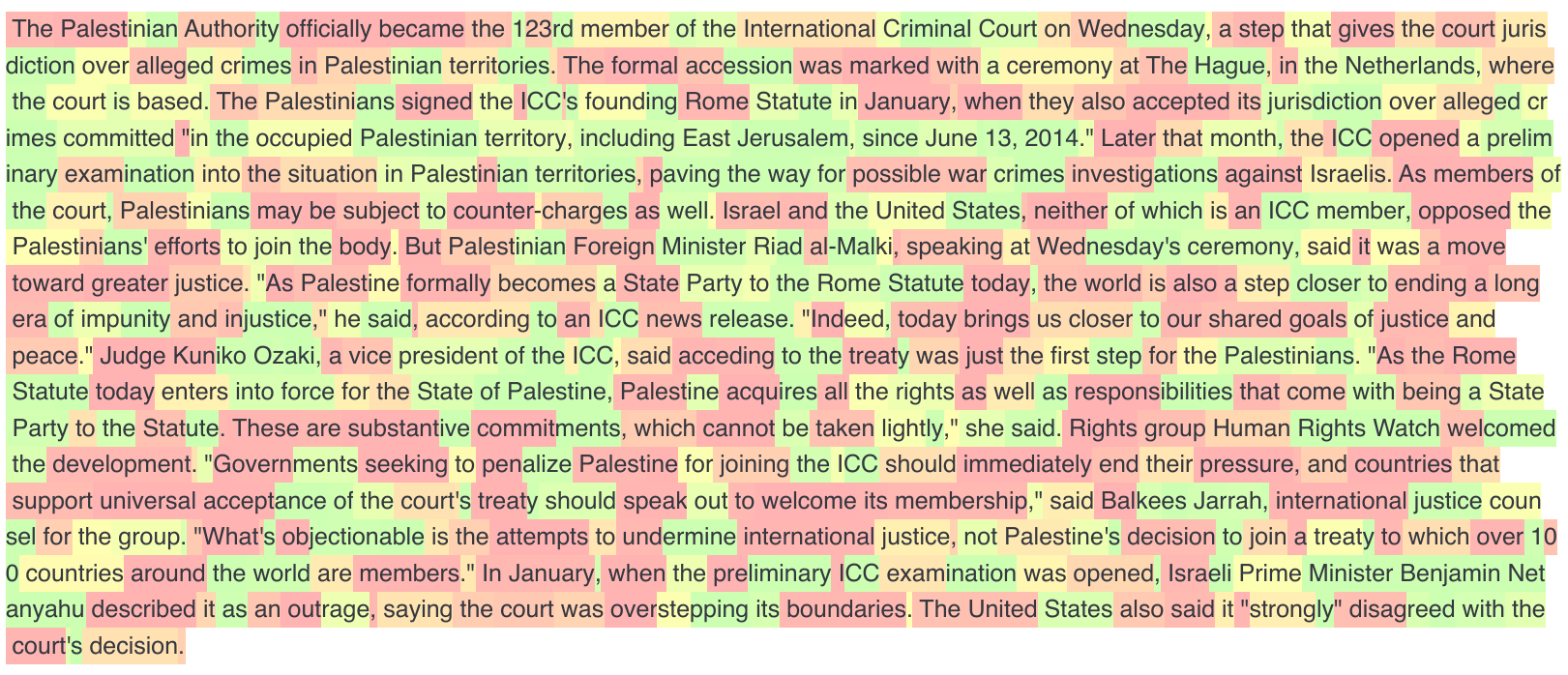}}
        \\
        \subfloat[\label{fig:case_2_b}Visualization of token probabilities estimated by LLaMA2-7b.]{
		\includegraphics[scale=0.58]{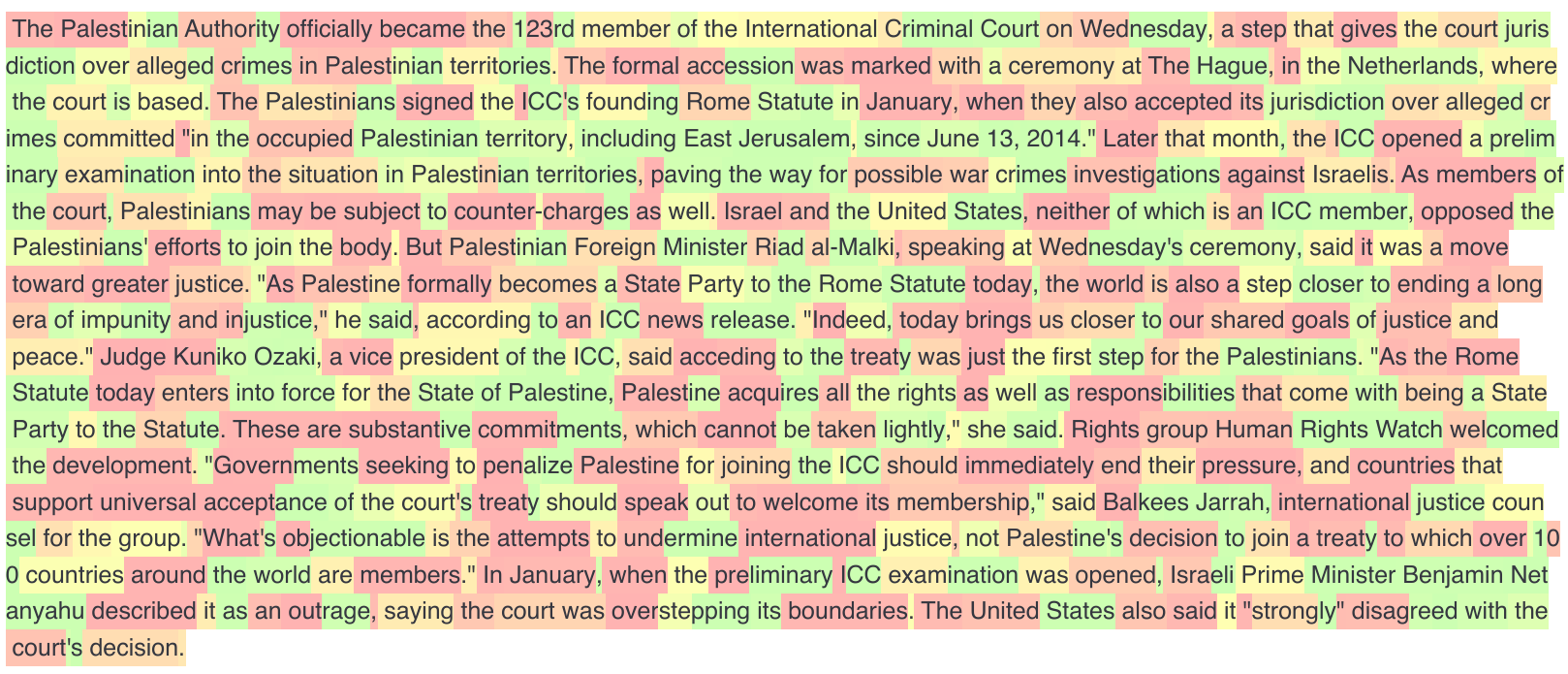}}
        \\
        \subfloat[\label{fig:case_2_c}Visualization of token probabilities estimated by Mistral-7b.]{
		\includegraphics[scale=0.58]{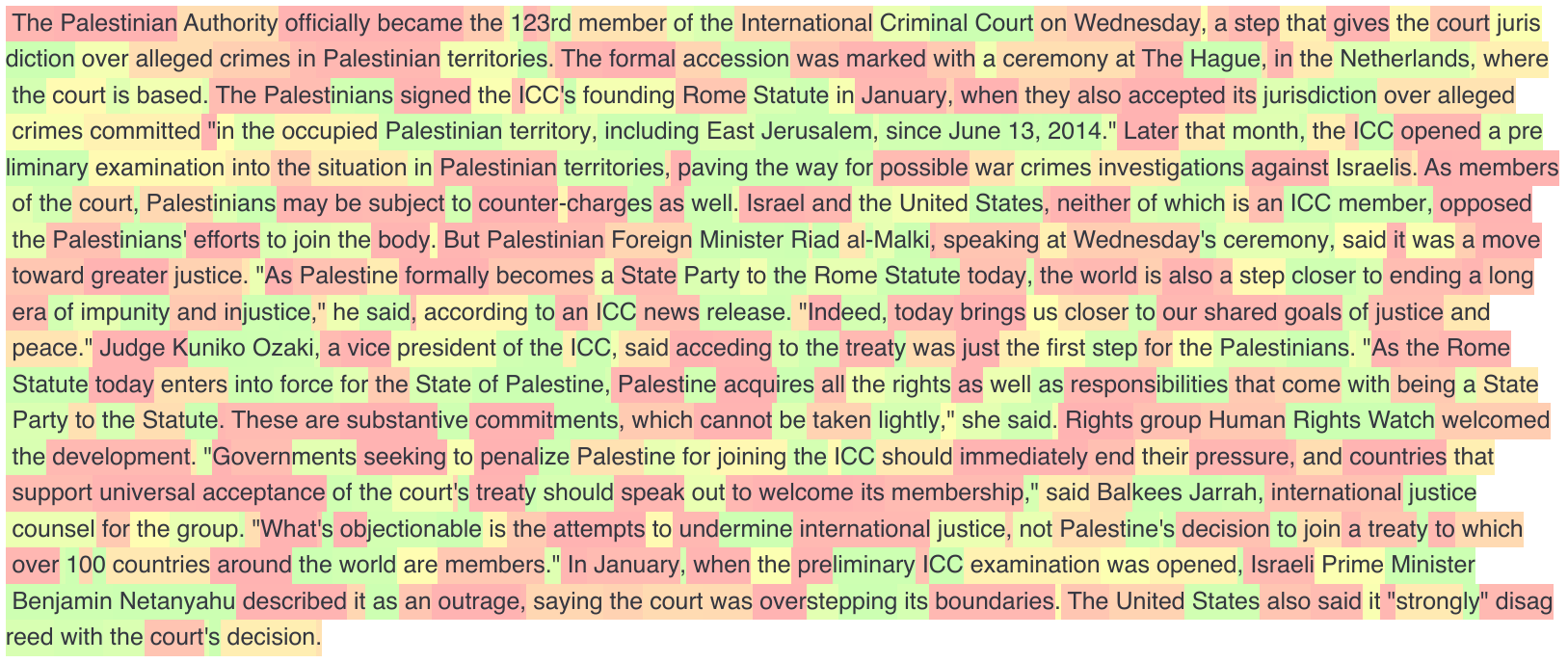}}
	\caption{Visualization of token probabilities estimated by LLaMA-7b, LLaMA2-7b and Mistral-7b on case 2. The color coding in the figure represents the token generation probabilities. A spectrum from red to green is utilized to represent a range from lower to higher generation probabilities, respectively.}
	\label{fig:case_2} 
\end{figure*}

\begin{figure*} [!t]
	\centering
	\subfloat[\label{fig:case_3_a}Visualization of token probabilities estimated by LLaMA-7b.]{
		\includegraphics[scale=0.58]{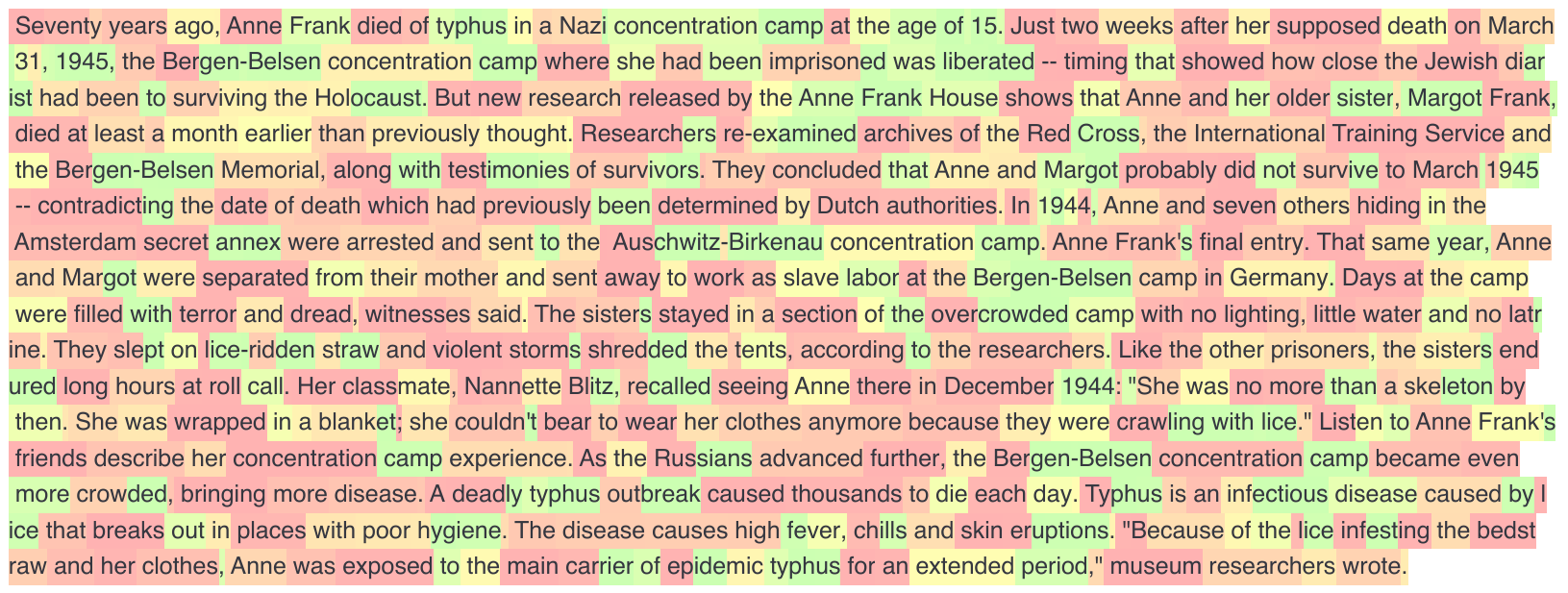}}
        \\
        \subfloat[\label{fig:case_3_b}Visualization of token probabilities estimated by LLaMA2-7b.]{
		\includegraphics[scale=0.58]{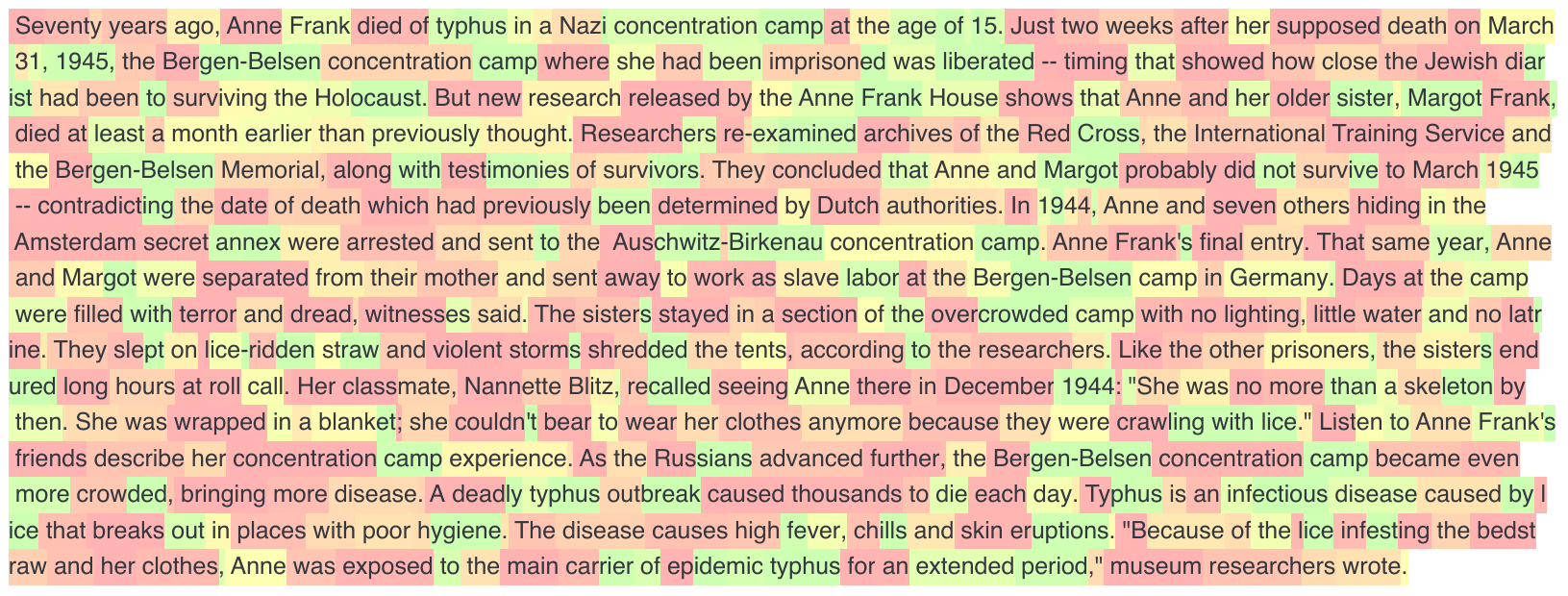}}
        \\
         \subfloat[\label{fig:case_3_c}Visualization of token probabilities estimated by Mistral-7b.]{
		\includegraphics[scale=0.58]{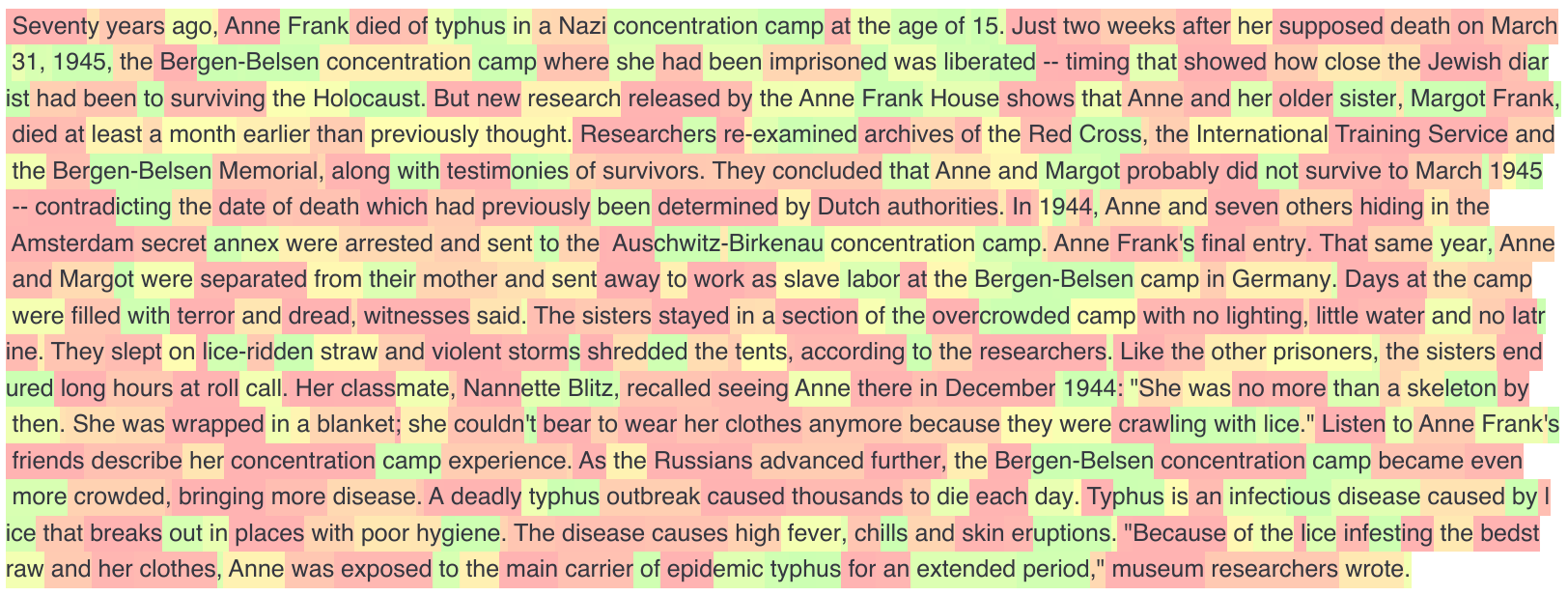}}
	\caption{Visualization of token probabilities estimated by LLaMA-7b, LLaMA2-7b and Mistral-7b on case 3. The color coding in the figure represents the token generation probabilities. Tokens with lower probabilities are colored more red, while tokens with higher probabilities are colored more green.}
	\label{fig:case_3} 
\end{figure*}

Figure~\ref{fig:case_1} to Figure~\ref{fig:case_3} illustrate three text examples where the background color of each token denotes its generation probability given by LLaMA-7b, LLaMA2-7b, and Mistral-7b. A spectrum from red to green is utilized to represent probabilities ranging from low to high. 

These examples reveal a tendency for lower probability tokens to consist of content words predominantly. However, instances exist where determiners and prepositions also exhibit lower probabilities, as exemplified in Figure~\ref{fig:case_1_a} with the words ``All'' and ``on'' in line 5, and the word ``in'' in line 14. This can be attributed to the interchangeable use of different words to convey identical semantic content at the beginning of a sentence or sub-clause. 

Besides, tokens with higher probabilities are typically from function words. However, exceptions are noted with certain content words demonstrating high probabilities due to the extensive knowledge memorized by LLMs, such as ``typhus'' and ``15'' in the first line of Figure~\ref{fig:case_3_a}. Additionally, some non-initial tokens, such as ``nesday'' and ``CC'' in lines 1 and 3 of Figure~\ref{fig:case_2_a}, exhibit high generation probabilities. It reflects the determinative role of initial tokens in setting the context for subsequent non-initial tokens. 

Comparing the prediction probabilities of LLaMA-7b, LLaMA2-7b, and Mistral-7b, we did not observe significant difference. Although LLaMA2-7b and Mistral-7b demonstrate superior performance in various benchmarks compared to LLaMA-7b, there is still considerable room for improvement in their grasp of factual knowledge. Therefore, it is beneficial to incorporate hesitations to make the model pay more attention to these informative key tokens.


\end{document}